\documentclass[10pt,journal,compsoc,twoside]{IEEEtran}

\makeatletter
\def\@IEEEsectpunct{.\ \,}
\def\paragraph{\@startsection{paragraph}{4}{\z@}{1\baselineskip plus 0.25\baselineskip minus 0.25\baselineskip}{0ex}{\normalfont\normalsize\bfseries}}
\makeatother

\usepackage[nocompress]{cite}

\usepackage[pdftex]{graphicx}

\usepackage{tikz}
\usepackage{pgfplots}
\usepackage{pgfplotstable}
\usepackage[eulergreek]{sansmath}

\usepackage{amsmath}
\usepackage{amssymb}
\usepackage{stmaryrd}

\usepackage{multirow}

\usepackage[colorlinks, allcolors={blue}]{hyperref}

\newcommand\addedSinceECCV[1]{#1}

\begin{document}

\title{\addedSinceECCV{Geometric Proxies} for Live RGB-D Stream \addedSinceECCV{Enhancement} and Consolidation}

\author{Adrien~Kaiser,~%
        Jose~Alonso~Ybanez~Zepeda,~%
        Tamy~Boubekeur%
\IEEEcompsocitemizethanks{%
\IEEEcompsocthanksitem A. Kaiser and J. A. Ybanez Zepeda are with Fogale Nanotech, 94270 Le Kremlin-Bicetre, France.%
\IEEEcompsocthanksitem A. Kaiser and T. Boubekeur are with Telecom Paris, 75013 Paris, France.%
}
}

\markboth{}{}

\IEEEtitleabstractindextext{
\begin{abstract}

We propose a \addedSinceECCV{geometric} superstructure for unified real-time processing of RGB-D data.
Modern RGB-D sensors are widely used for indoor 3D capture, with applications ranging from modeling to robotics, through augmented reality.
Nevertheless, their use is limited by their low resolution, with frames often corrupted with noise, missing data and temporal inconsistencies.
Our approach consists in generating and updating through time a single set of compact local statistics parameterized over detected \addedSinceECCV{geometric} proxies, which are fed from raw RGB-D data.
Our proxies provide several processing primitives, which improve the quality of the RGB-D stream on the fly or lighten further operations.
Experimental results confirm that our lightweight analysis framework copes well with embedded execution as well as moderate memory and computational capabilities compared to state-of-the-art methods.
Processing RGB-D data with our proxies \addedSinceECCV{allows} noise and temporal flickering removal, hole filling and resampling.
As a substitute of the observed scene, our proxies can additionally be applied to compression and scene reconstruction.
We present experiments performed with our framework in indoor scenes of different natures within a recent open RGB-D dataset.
\end{abstract}

}

\maketitle

\IEEEraisesectionheading{\section{Introduction}\label{sec:introduction}}

\IEEEPARstart{T}{he} real time RGB-D stream output of modern commodity consumer depth cameras can feed a growing set of end applications,
from human computer interaction and augmented reality to industrial design.
Although such devices are constantly improving, the limited quality of their stream still restraints their impact spectrum.
This mostly originates in the low resolution of the frames and the inherent noise, incompleteness and temporal inconsistency stemming from single view capture.

\label{sec:introduction_objectives}

\begin{figure*} %
\centering
\includegraphics[width=\textwidth]{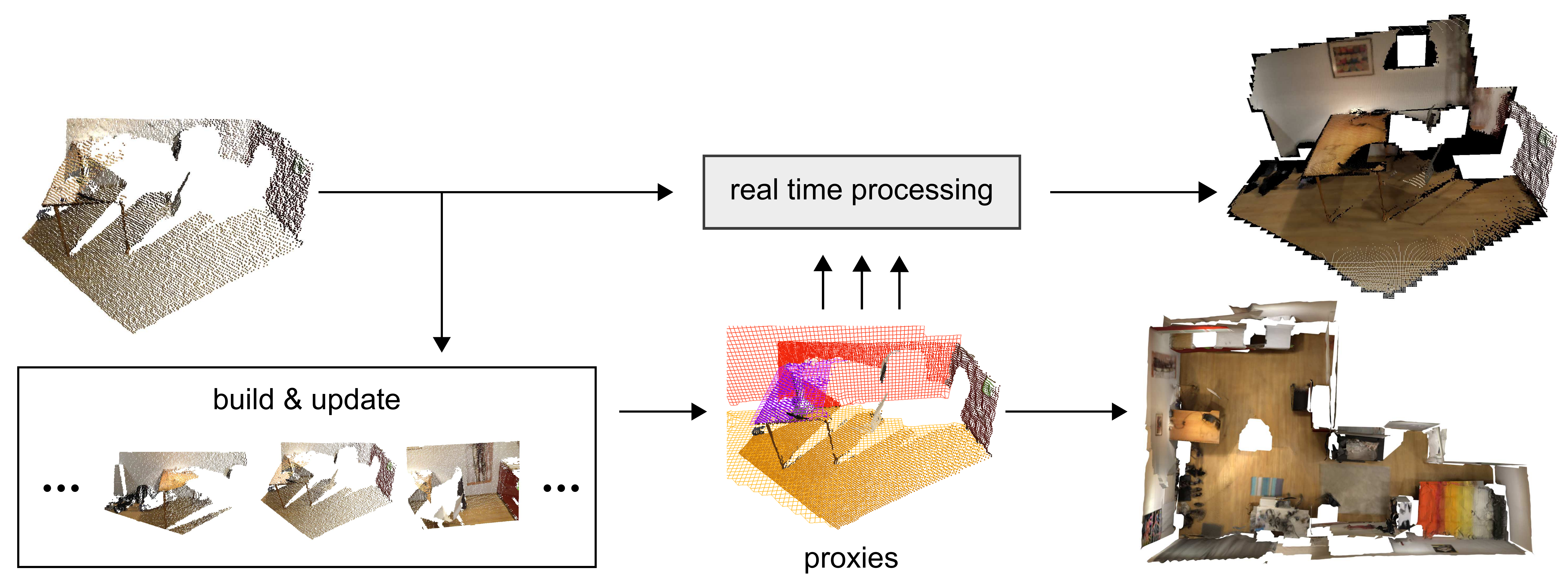}
\caption{\emph{Overview} \addedSinceECCV{of our geometric proxy framework.}
From a stream of \addedSinceECCV{2.5D} RGB-D frames (left),
\emph{proxies} are built on the fly and updated over time (bottom, detailed in \autoref{fig:proxies_building})
and used to apply different real-time processing primitives to the incoming RGB-D frames (top).
The system outputs an enhanced data stream and a \addedSinceECCV{geometric regular} model of the observed scene (right).}
\label{fig:introduction_overview}
\end{figure*}

\addedSinceECCV{We introduce a new multi-shape geometric superstructure to} improve RGB-D streams \addedSinceECCV{on the fly} by analyzing them.
A sparse set of detected 3D \addedSinceECCV{primitive shapes is} parameterized to record statistics extracted from the stream and form\addedSinceECCV{s} a structure that we call \emph{proxies}.
This superstructure substitutes the RGB-D data and approximates the geometry of the scene.
\addedSinceECCV{We define a geometric framework using the time-evolving statistics stored in our proxies
and based on their consistent spatial support that can be then seen as geometric "scaffolding". %
Its purpose is to} improve the RGB-D stream on the fly by reinforcing features, removing noise and outliers or filling missing parts,
under the memory-limited and real time embedded constraints of mobile capture in indoor environments (\autoref{fig:introduction_overview}).

We designed such a lightweight \addedSinceECCV{geometric} superstructure to be stable through time and space,
which gives priors to apply several signal-inspired processing primitives to the RGB-D frames.
They include filtering to remove noise and temporal flickering, hole filling or resampling (\autoref{sec:processing}).
This allows structuring the data and \addedSinceECCV{can} simplify or lighten subsequent operations, e.g. tracking and mapping, \addedSinceECCV{automated navigation}, measurement, data transmission, rendering or physical simulation.
While our primary goal is the enhancement of the RGB-D data stream, our framework can additionally be applied to compression (\autoref{sec:experiments_processing_compression})
and scene reconstruction (\autoref{sec:consolidation}), as the \addedSinceECCV{proxy} structure is a representation of the observed scene.

\label{sec:introduction_overview}

In practice, our system takes a raw RGB-D stream as input to build and update a set of \addedSinceECCV{geometric} proxies on the fly.
It outputs an enhanced stream together with a reconstruction of \addedSinceECCV{regularly-shaped (e.g., planar)} areas in the observed scene. %
\addedSinceECCV{A selection of proxies based on the current RGB-D frame can be used for lightened transmission of the data.
Proxies can be used as priors for triangulation and fast depth data meshing, with applications to rendering or simulation.}
On the contrary to previous approaches, which mostly rely on a full volumetric reconstruction to consolidate data,
our approach is lightweight, with a moderate memory footprint and a transparent interfacing to any higher level RGB-D pipeline.

\addedSinceECCV{In particular}, our contributions are:

\begin{itemize}
    \item a stable and lightweight \addedSinceECCV{geometric} superstructure for RGB-D data, \addedSinceECCV{unified for multiple shapes} (\autoref{sec:proxies_building_extent});
    \item construction and updating methods which are spatially and temporally consistent (\autoref{sec:proxies_building_shapes});
    \item \addedSinceECCV{a set of compact statistics that record local information from RGB-D samples (\autoref{sec:proxies_statistics});}
    \item a collection of RGB-D enhancement methods based on our structure (\autoref{sec:processing}) which run on the fly \addedSinceECCV{to generate a global scene reconstruction (\autoref{sec:consolidation})}.
\end{itemize}

\addedSinceECCV{This paper extends our preliminary work \cite{kaiser2018proxy}, including:} %

\begin{itemize}
\addedSinceECCV{ %
    \item generalization to cylinders and spheres (\autoref{sec:proxies_building_extent})
    \item additional cell-wise color statistics (\autoref{sec:proxies_statistics_local})
    \item new experiments on synthetic scenes (\autoref{sec:experiments_synthetic}). %
}
\end{itemize}

\section{Related Work}
\label{sec:previouswork}

\subsection{RGB-D Data Improvement}
\label{sec:previouswork_improvement}

\paragraph*{Filtering}

Depth maps can be denoised using spatial filters e.g., Gaussian, median, bilateral \cite{tomasi1998bilateral},
adaptive or anisotropic \cite{liu2016acomputationally} filters, often refined through time,
with the resulting enhanced stream potentially used for a full 3D reconstruction \cite{newcombe2011kinectfusion}.
Other methods include non-local means, bilateral filters with time, %
Kalman filters, over-segmentation and region growing. %
Wu et al. \cite{wu2014realtime} present a \emph{shape-from-shading} method using the color component to improve the geometry,
which allows adding details to the low quality input depth.
They show applications of their method to improve volumetric reconstruction on multiple small scale and close range scenes.

Depth maps can be upsampled using cross bilateral filters such as \emph{joint bilateral upsampling} \cite{kopf2007joint} or \emph{weighted mode filtering} \cite{min2012depth}.
Such methods are particularly useful to recover sharp depth regions boundaries and enforce depth-based segmentation.

\paragraph*{Hole Filling}

Range limits and high noise levels of depth sensing often create holes in RGB-D data.
Given the material of observed objects and the type of technology used, e.g. \emph{time of flight}, \emph{light coding} or \emph{stereo vision}, some surfaces are harder to detect.
The orientation of the surface with regards to the sensor and the perturbations due to light sources can also lower the quality in certain areas.
In order to fill these holes in the depth component, one can use the same spatial filters as those used for denoising, %
or morphological filters \cite{liu2016acomputationally}.
Inpainting methods \cite{liu2016hole}, over-segmentation or multi-scale processing %
are also used to fill holes for e.g., \emph{depth image-based rendering} (DIBR) under close viewing conditions.

\paragraph*{Shape-based Depth Processing}

A set of 3D \addedSinceECCV{shapes} offers a faithful yet lightweight approximation for many indoor environments. 
Surprisingly, only a few methods have used shape proxies as priors to process 2.5D data,
with in particular Schnabel et al. \cite{schnabel2009completion} who detect limits of shapes to fill in holes in static 3D point clouds.
\emph{Fast sampling plane filtering} \cite{biswas2012planar} detects and merges planar patches in static indoor scenes. The detected planes allow filtering the planar surfaces of the input point cloud, however the primitives seem quite sensitive to the depth sensor noise and lack spatial consistency.

\subsection{Simple Primitive Shape Detection}
\label{sec:previouswork_simpleshapes}

Methods that build high level models of captured 3D data, \addedSinceECCV{composed of simple geometric primitive shapes,} are mostly based on \emph{RANSAC} \cite{fischler1981random}, the \emph{Hough transform} \cite{hulik2014continuous} or \emph{Region Growing} algorithms.
In our embedded, real time, memory-limited context, we take inspiration from the RANSAC-based method of Schnabel et al. \cite{schnabel2007efficient}
for its time and memory efficiency,
by repeating \addedSinceECCV{shape} detection through time to acquire a consistent model and cope with the stochastic nature of RANSAC.

Their \emph{efficient RANSAC} implementation gives stochastic improvements to the critical steps of the algorithm in terms of complexity.
For a regular RANSAC-based \addedSinceECCV{shape} detection, minimal sets of three \addedSinceECCV{oriented} points would be randomly picked a fixed and large number of times.
Then, the shape parameters are estimated from this minimal set and inliers of the estimated shape are computed.
The shape with the highest score is kept, its inliers are removed from the point cloud and the algorithm is ran again on the remaining data.
Schnabel et al. replace the fixed number of loops with a stochastic condition to stop looking for \addedSinceECCV{shapes} in the dataset, based on the number of detected shapes and number of randomly picked minimal sets.
In addition, instead of searching the full point cloud for inliers of a given shape, they estimate this count in a random subset of the dataset and extrapolate it to the full point cloud.
Other modifications allow improving the quality of detected shapes with a localized sampling and specific post-processing.

Our framework is not attached to a particular \addedSinceECCV{shape} detection method, and other algorithms such as \emph{point clustering} \cite{holz2011realtime}
or \emph{agglomerative hierarchical clustering} \cite{feng2014fastplane}, could be used.
For a complete overview of methods to detect simple \addedSinceECCV{geometric primitive} shapes in captured 3D data, we refer the reader to a recent survey \cite{kaiser2018survey}.

\subsection{RGB-D Stream Reconstruction} %
\label{sec:previouswork_reconstruction}

\paragraph*{Dense SLAM}

Online dense \emph{simultaneous localization and mapping (SLAM)} methods accumulate points within a map of the environment, while continuously localizing the sensor in this map.
Recent dense SLAM systems include \emph{KDP SLAM} \cite{hsiao2017keyframe} or \emph{ORB-SLAM2} \cite{muratal2017orbslam2}.
Point-based fusion \cite{keller2013realtime} is also used to accumulate points without the need of a full volumetric representation.

\addedSinceECCV{In our implementation, we use \emph{RGB-D SLAM} \cite{endres20143dmapping} which leverages} point features detected in the color component of the RGB-D frame to estimate camera motion.
After detecting and matching SIFT, SURF or ORB features in subsequent color images, their 3D positions in both frames are computed using the depth component.
Using these matching 3D points, a robust RANSAC-based estimation of the motion matrix allows discarding false positive matches.
Sets of three matching points are randomly picked and the matrix transforming a set in the first frame into the second set is computed using a \emph{least squares} method. %
Inliers of the transformation are estimated using their 3D position and orientation and the one giving the most inliers is kept.
It is important to note that any existing method or device that localizes an RGB-D camera in its environment can be used instead.

\paragraph*{Volumetric Depth Fusion}

\addedSinceECCV{\emph{Depth fusion} builds a volumetric representation of an input scene by accumulating depth observations into a voxel grid to update values of signed distance to the nearest model surface.}
Scene reconstruction methods using volumetric fusion \addedSinceECCV{became popular with} \emph{KinectFusion} \cite{newcombe2011kinectfusion}, \addedSinceECCV{as the first online reconstruction method based on consumer grade depth sensor \emph{Kinect}.
Recent optimizations include} \emph{VoxelHashing} \cite{niessner2013realtime} for efficiency and \emph{BundleFusion} \cite{dai2017bundle} for accuracy.
However, the need for a voxel grid representing the space leads to high requirements of memory.

\paragraph*{Offline Surface Regularization}

Several methods have been developed to include \addedSinceECCV{geometric} primitives in the SLAM system, either to smooth and improve the reconstruction \cite{salasmoreno2014dense} or improve the localization of the sensor \cite{kaess2015simultaneous}.
A recent offline method \cite{zhang2016emptying} makes use of planes to estimate the geometry of a room in order to remove furnitures and model the lighting of the environment.
This allows the user to re-light and re-furnish the room as desired.
Some recent algorithms make use of planes to smooth and complete the data within a depth fusion volume, such as methods by Zhang et al. \cite{zhang2015online} or Dzitsiuk et al. \cite{dzitsiuk2017denoising}.
Offline improvement methods have been developed based on the volumetric representation of the scene,
such as \emph{3DLite} \cite{huang20173dlite} that builds a planar model of the observed scene and optimizes it to achieve a high quality texturing of the surfaces.

\section{Geometric Proxies}
\label{sec:proxies}

\begin{figure*} %
\centering
    \includegraphics[width=\textwidth]{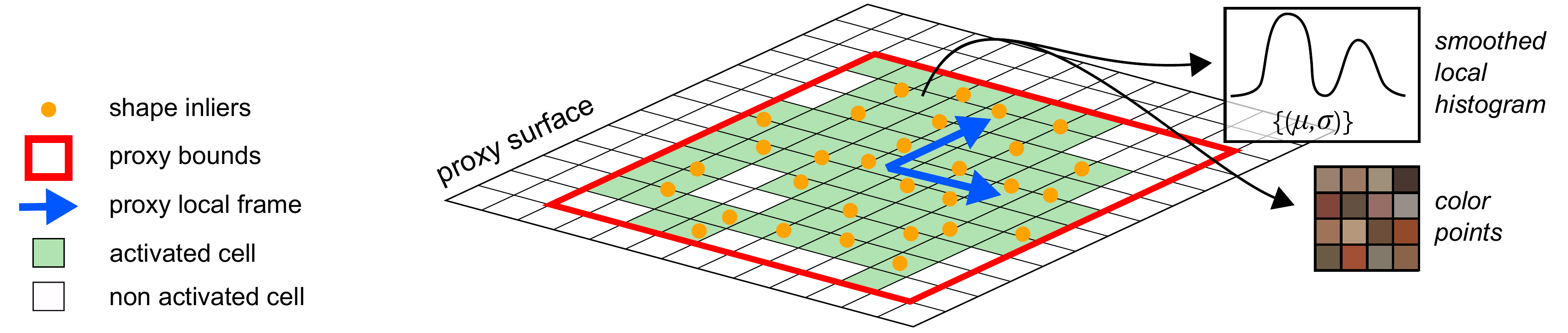}
\caption{Proxy model.
Built upon a \addedSinceECCV{shape} in 3D space \addedSinceECCV{(e.g., a plane)}, our proxy model is made of a local frame, bounds and a grid of cells containing statistics.
These statistics are a collection of mean $\mu$ and variance $\sigma$ values representing a \emph{smoothed local histogram},
\addedSinceECCV{as well as quantized color information}, all gathered from the RGB-D data (\autoref{sec:proxies_statistics}).
Activated cells are the ones containing inliers from many frames.}
\label{fig:proxies_model}
\end{figure*}

Basically, \addedSinceECCV{our model represents} RGB-D data which is often seen and consistent through frames and space, hence revealing the dominant structural elements in the scene.
To do so, it takes the form of a \addedSinceECCV{geometric} superstructure, \addedSinceECCV{made of multiple shape proxies}, all equipped with a local frame, bounds and, within the bounds,
a regular 2D grid of rich statistics, mapped on the \addedSinceECCV{shape} and gathered from the RGB-D data.
\autoref{fig:proxies_model} gives visual insight of a \addedSinceECCV{geometric} proxy.
\addedSinceECCV{A proxy can have the shape of a plane, a cylinder or a sphere.
Our implementation is based on the \emph{efficient RANSAC} shape detection method by Schnabel et al. \cite{schnabel2007efficient}.}

\subsection{Building Geometric Proxies}
\label{sec:proxies_building}

\subsubsection{Shape Detection and Tracking}
\label{sec:proxies_building_shapes}

\begin{figure*} %
\centering
    \includegraphics[width=\textwidth]{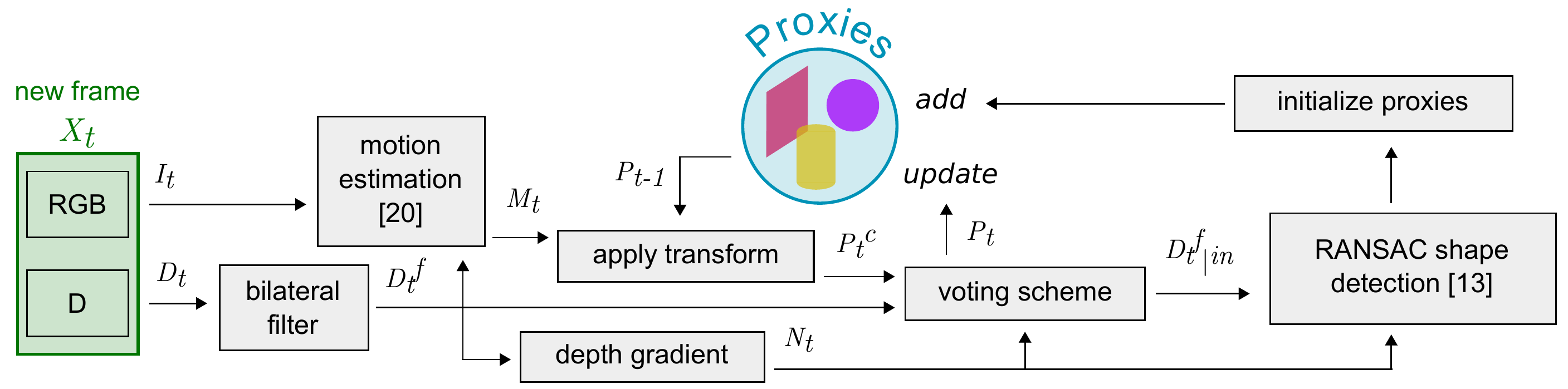}
\caption{\addedSinceECCV{\emph{Overview} of proxy structure construction.}
This procedure is ran for each new RGB-D image $X_t$ at frame $t$, made of a color image $I_t$ and a depth map $D_t$.
\addedSinceECCV{The proxy update allows \emph{accumulating} samples from $D_t$ in the \emph{superstructure} to compute statistics, as detailed in \autoref{sec:proxies_statistics}.}
$D_t^f$: low-pass filtered version of $D_t$;
$M_t$: camera motion matrix;
$N_t$: normal vectors associated with $D_t$;
$P_{t-1}$: proxies detected at frame $t-1$;
$P_t^c$: candidate proxies; $P_t$: proxies at frame $t$;
$D_t^f \scriptstyle |in$: low-pass filtered depth points without inliers from $P_t$.}
\label{fig:proxies_building}
\end{figure*}

\begin{figure*} %
\centering
    \includegraphics[width=\textwidth]{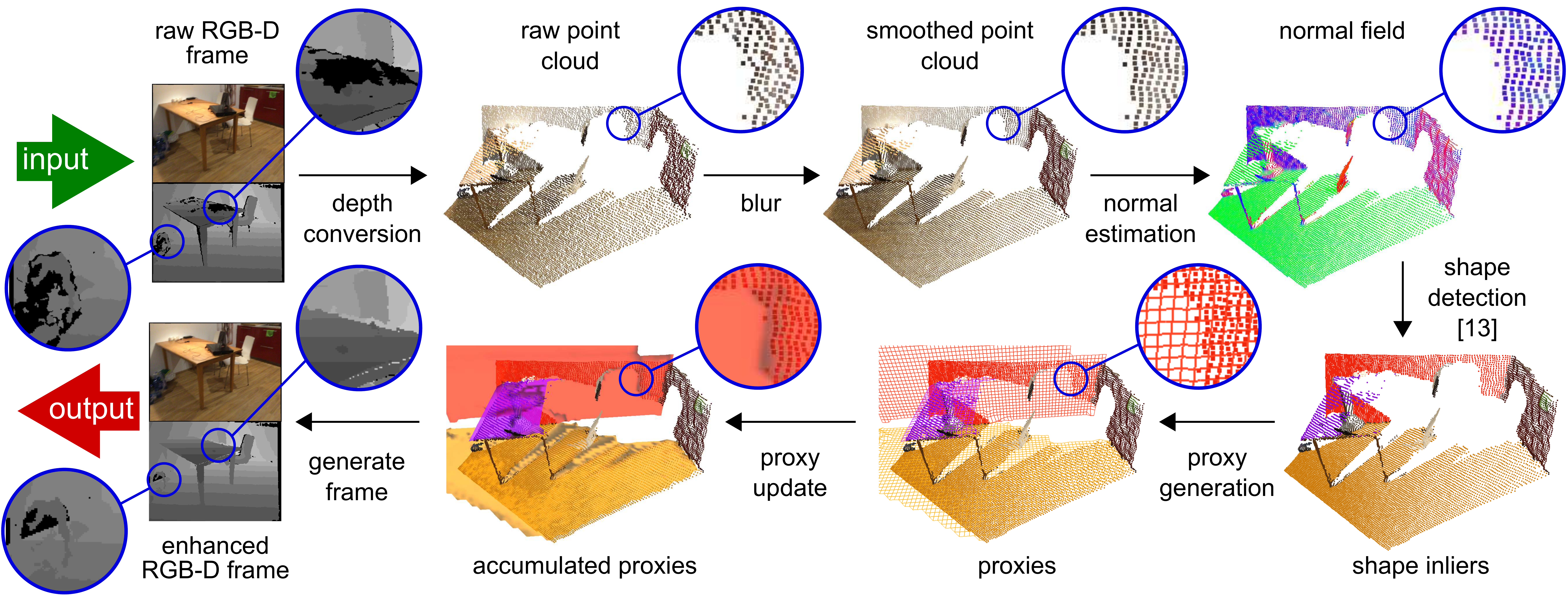}
\caption{\addedSinceECCV{Example of proxy structure construction.}
Top: the input RGB-D frame is converted into a raw point cloud to which a low-pass filter is applied, followed by normal estimation.
Right: RANSAC-based \addedSinceECCV{shape} detection \cite{schnabel2007efficient} is applied and used as a basis for the construction or the update of our proxies,
following the structure shown in \autoref{fig:proxies_model}.
Bottom: accumulated proxy cells are visualized with colors weighted by their occupancy probabilities:
darker cells have a low probability and represent low confidence areas, whereas brighter cells represent areas with high confidence.
Proxies are then used to generate enhanced RGB-D frames, as detailed in \autoref{sec:processing}. The whole process runs on the fly.}
\label{fig:proxies_buildingexample}
\end{figure*}

We build \addedSinceECCV{geometric} proxies on the fly and update them through time using solely incoming raw RGB-D frames from the live stream.
More precisely, for each new RGB-D image $X_t = \{I_t , D_t\}$ (color and depth),
we run the procedure described in \autoref{fig:proxies_building}
and shown in \autoref{fig:proxies_buildingexample} on one specific example.

The initial depth filtering is based on a bilateral convolution \cite{tomasi1998bilateral} of the depth map using a Gaussian kernel
associated with a range check. \addedSinceECCV{This allows} discarding points further than a depth threshold from the current point, which could create artificial depth values if taken into account.
In our experiments, we choose to set this threshold to \emph{20cm}, which allows filtering together parts of the same object, while ignoring the influence of unrelated objects.
We then estimate the normal field through the simple computation of the depth gradient at each pixel, due to the embedded processing constraint, using the sensor topology as domain.

The estimation of the camera motion from the previous frame is inspired from \addedSinceECCV{\emph{RGB-D SLAM}} introduced by Endres et al. \cite{endres20143dmapping}, using point features from $I_t$.
\addedSinceECCV{However}, any egomotion estimation algorithm can be used at this step, as all we need is the values of the six degrees of freedom \addedSinceECCV{localizing an RGB-D camera in its environment.}
Examples of such algorithms are given in \autoref{sec:previouswork_reconstruction}.

In order to keep or discard previously detected proxies, we define a voting scheme where samples of $X_t$ which are inliers of a given previous proxy cast their vote to this proxy and are marked.
Then, the per-proxy vote count \addedSinceECCV{in the new frame} indicates whether the proxy is preserved or discarded.
Preserved proxies are updated with $X_t$, hence see their parameters refined and occupancy statistics updated with new inliers.
Discarded proxies are placed in \emph{probation} state for near-future recheck with new incoming frames, and purged if discarded for too long.
However, in order to avoid losing information on non-observed \addedSinceECCV{but important} parts of the scene, we do not purge proxies that have been seen \addedSinceECCV{a large number of times}, which stay in probation instead.

When new proxies have been detected, similar ones are merged together in order to avoid modeling different parts of geometric surfaces with multiple proxy instances.
The proxy is then \addedSinceECCV{initialized} with a bounding rectangle and a local frame computed to be aligned with the scene orientation, \addedSinceECCV{based on the \emph{Manhattan world} assumption \cite{coughlan1999manhattan}} (more details in the supplemental material, \autoref{S_sec:experiments_implementation_orientation}). %
Using the global scene axes to compute the local \addedSinceECCV{proxy} frame leads to a fixed resolution and spatial consistency for the grid of all proxies and allows efficient recovery and fusion.

Finally, \addedSinceECCV{initial} occupancy statistics are \addedSinceECCV{computed} using $X_t$,
\addedSinceECCV{by recovering the coordinates of the cell in the proxy grid corresponding to each inlier.}
In order to take into account the point of view when modeling the scene, \addedSinceECCV{grid cell coordinates for an inlier depth point are computed by projecting it} upon the detected shape following the direction between the camera and the point.
\addedSinceECCV{This projection is illustrated in \autoref{S_fig:pointfiltering} of the supplemental material.} %
As a last step, proxies are transformed from local depth frame into global 3D space in order to be tracked in the next frames.

\subsubsection{Spatial Extent Modeling}
\label{sec:proxies_building_extent}

\addedSinceECCV{

In order to model the observed elements as faithfully as possible, we need to take into account the extent of these elements at the surface of the shape.
In addition, we wish to store local information on parts of the shape to avoid smoothing small details,
which requires quantization of the shape surface.
When proxy shapes are detected, the shape is parameterized based on its local frame in order to define a consistent grid, fixed with relation to the actual object.
In particular, densely modeling local geometry implies requirements of minimum cell distortion for uniform representation of local information and efficient unfolding to 2D space.
The parameterization from 3D space at the surface of the shape to 2D space is described below for planes and revolution surfaces such as cylinders and spheres.

\paragraph*{Planes}

We define local axes in the space of a plane based on the reference orientation of the world, as detailed in the supplemental material (\autoref{S_sec:experiments_implementation_orientation}). %
This leads to a consistent orientation of the grid of all proxies, where local proxy axes $\vec X$ and $\vec Y$ both belong to the plane.
This local frame allows us to compute a consistent parameterization of the shape for any input image.
In particular, a 3D point $P$ belonging to a plane of origin point $C$ will have coordinates in the local frame of the proxy defined as

} %

\begin{equation}
\addedSinceECCV{ %
\left\{
\begin{array}{l}
u = (P - C) . \vec X\\
v = (P - C) . \vec Y ~ .\\
\end{array}
\right. \\
} %
\label{eq:proxies_parameterization_plane}
\end{equation}

\addedSinceECCV{

\paragraph*{Revolution Shapes}

For the shapes of revolution such as cylinders and spheres, we want to maintain a unified representation allowing the use of the same 2D operators as for the plane.
In that regard, we define a parameterization for both of these shapes, also based on their local axes.
\autoref{fig:proxies_parameterization} shows the parameterization of the revolution shapes.

\begin{figure*} %
\centering
    \includegraphics[width=\textwidth]{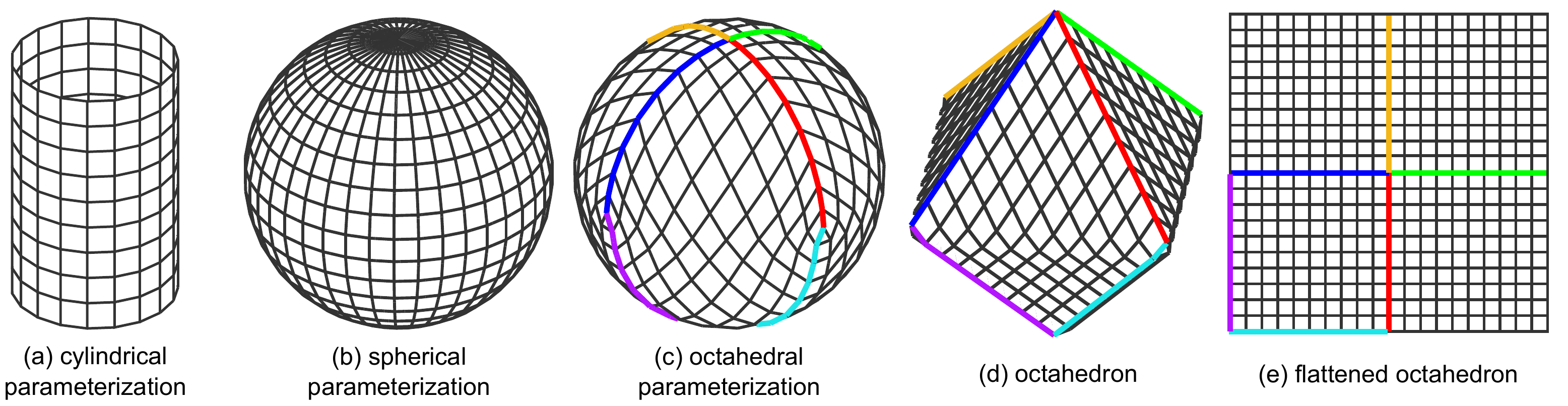}
\caption{\addedSinceECCV{Proxy grid parameterization for cylinders and spheres.
Cylinders are parameterized with the cylindrical coordinates (a).
As parameterizing a sphere with spherical coordinates creates heavy distortion at the poles (b),
we choose to use the octahedral parameterization (c), which reduces stretch between grid cells.
This parameterization defined by Praun and Hoppe \cite{praun2003spherical} uses an octahedron (d) to reduce stretch at the surface of the sphere.
The corresponding 2D grid (e) fully contains the extent of the 3D shape.}}
\label{fig:proxies_parameterization}
\end{figure*}

The cylinder proxy has its $\vec X$ and $\vec Y$ local axes orthogonal to its direction axis $\vec A$, allowing the use of \emph{cylindrical coordinates}.
Hence, the local axes define the \emph{angular position} around the cylinder, while the axis $\vec A$ defines the \emph{height} of a point along the cylinder.
On a cylinder of origin point $C$ and radius $r$, a 3D point $P$ would have its local shape coordinates defined as

} %

\begin{equation}
\addedSinceECCV{ %
\begin{array}{l}
\left\{
\begin{array}{l}
u = r (\pi + \text{arctan2}(\vec pc . \vec Y, \vec pc . \vec X)) \in [0, 2 \pi r]\\ %
v = \vec pc . \vec A\\
\end{array}
\right. \\
\text{with } \vec pc = P - C ~ .\\
\end{array}
} %
\label{eq:sproxies_parameterization_cylinder}
\end{equation}

\addedSinceECCV{

For the sphere, the straightforward parameterization would be based on spherical coordinates.
Local axes $\vec X$ and $\vec Y$ of the sphere proxy are defined in that sense, where they allow computing the \emph{azimuthal} angle.
In that parameterization, the \emph{polar} angle is computed from the \emph{zenith} direction defined as $\vec X \times \vec Y$.
However, as we can see in \autoref{fig:proxies_parameterization} (b), the use of spherical coordinates implies strong distortion at the poles.

As we want to discretize the shape extent to locally represent the surface with statistics, it would be more meaningful to have a grid with similar cell sizes,
while keeping a fast and efficient unfolding to 2D space.
In consequence, we choose to parameterize a sphere with an \emph{octahedron}, as defined by Praun and Hoppe \cite{praun2003spherical}, which strongly reduces stretch and conveniently unfolds to a square with no space lost.

On a sphere of center point $C$ and radius $r$ parameterized as an octahedron, a 3D point $P$ would have its local shape coordinates defined as

} %

\begin{equation}
\addedSinceECCV{ %
\begin{array}{l}
\left\{
\begin{array}{l}
u = \frac{\pi r}{2} \left\{
\begin{array}{l l}
\frac{x}{n} & \text{if } z \geq 0 \\
1 - \frac{|y|}{n} & \text{if } z < 0 \wedge x \geq 0 \\
\frac{|y|}{n} - 1 & \text{if } z < 0 \wedge x < 0 \\
\end{array}
\right. \\
v = \frac{\pi r}{2} \left\{
\begin{array}{l l}
\frac{y}{n} & \text{if } z \geq 0 \\
1 - \frac{|x|}{n} & \text{if } z < 0 \wedge y \geq 0 \\
\frac{|x|}{n} - 1 & \text{if } z < 0 \wedge y < 0 \\
\end{array}
\right. \\
\end{array}
\right. \\
\text{with } \left\{
\begin{array}{l}
x = \vec pc . \vec X \\
y = \vec pc . \vec Y \\
z = \vec pc . (\vec X \times \vec Y) \\
\end{array}
\right. 
\text{with } \vec pc = \frac{P - C}{\|P - C\|} \\
\text{and } n = |x| + |y| + |z| ~ . \\
\end{array}
} %
\label{eq:proxies_parameterization_sphere}
\end{equation}

\paragraph*{Extent Discretization}

\addedSinceECCV{From the 2D local coordinates of each shape as described above, we define a fixed-size grid on top of the shape surface.
We simply discretize the coordinate values \addedSinceECCV{$(u, v)$} of a given point at the surface of shape using a fixed cell size.
We set the} size to \emph{5cm} x \emph{5cm}, which corresponds to about four times the area of a depth pixel at a typical distance of 8 meters \addedSinceECCV{from the sensor}.
The area of a pixel at given depth $z$ is given by $a(z) = \tan(\frac{fov_H}{res_H})  \tan(\frac{fov_V}{res_V})  z^2$.
With $fov= (60^{\circ},45^{\circ})$ and $res=(320,240)$, we have $a(8m) \approx 0.00068539m^2 \approx (2.6cm)^2 \addedSinceECCV{\approx \frac{(5cm)^2}{4}}$.
Hence, this size ensures a minimum sampling of proxy cells by depth points even at far capture distances.
\addedSinceECCV{In practice, the real case capture distance will not go beyond 5 to 6 meters due to the size of indoor rooms and limitations of the capture device.
In consequence, the potential amount of visiting depth point per frame for a cell is guaranteed to be more than four.}

Cells are activated when their visitation percentage over the recent frames (the last 100 frames in our experiments) is greater than a threshold (25\% in practice).
Once activated, a cell stays so until the end of the processing.
We consider a cell as visited as soon as it admits one inlier data point i.e., a data point located within a threshold distance to the cell under a projection in the direction from the sensor origin to the point.
This activation threshold allows modeling the actual geometry of the observed scene, while discarding outliers observations due to the low quality of the sensor.

\subsection{Proxy Statistics}
\label{sec:proxies_statistics}

\subsubsection{Accumulation of Depth Samples}
\label{sec:proxies_statistics_accumulation}

\addedSinceECCV{

As shown in \autoref{sec:proxies_building_shapes}, we track 3D shapes in time and space to maintain a consistent geometric representation of the surroundings.
This temporal and spatial consistency gives information on local geometry at its fixed location in the real world.
Hence, we leverage the 3D sampling of this geometry as provided by the depth sensors, to refine our model with time.
At each new frame, shape tracking provides a list of 3D sample points belonging to a shape.
These 3D points bear geometric and appearance information such as distance to the shape, orientation deviation, curvature and color.
The definition of our local extent model at the shape surface, as explained in \autoref{sec:proxies_building_extent},
allows accumulating this information into a unified geometric representation in shape space.

The first use of these accumulated depth samples at the shape surface is to compute global shape statistics.
At each frame, the new set of inliers is used to compute shape parameters which are averaged with the previous parameters to get more consistent shapes.
In addition, the standard deviation of shape parameters gives insight on their distribution and can be seen as a \emph{confidence} value on the shape.

}

\subsubsection{Local Statistics}
\label{sec:proxies_statistics_local}

\addedSinceECCV{

The consistent grid defined at the surface of a shape based on a shape-specific parameterization was designed to gather statistics on local parts of the shape,
in order to model small details.
In each cell of the grid, a probability of \emph{occupancy} is computed to get knowledge of the visitation rate of this cell.
This allows understanding whether a cell models actual geometry in the scene, or if it just corresponds to noise or flickering parts and should be ignored.
A real data example of statistics of occupancy and distance stored at the surface of shapes is shown as supplemental material (\autoref{S_fig:statistics_image}).

}

\paragraph*{Shape Distance}

Each cell of the grid includes a statistical model of the depth values \addedSinceECCV{gathered from the RGB-D data.
These statistics are composed of a collection of mean $\mu$ and variance $\sigma$ values of the distance to the shape.
This local distribution} represents a \emph{smoothed local histogram} \cite{kass2010smoothed} made of Gaussian kernels.
\addedSinceECCV{Using such a compressed histogram representation allows recording samples into a compact but faithful model
made of a short list of normal distribution parameters, as well as smoothing out outlier depth values.}
The contribution of an inlier $p$ of distance $d(p)$ to the proxy is given as

\begin{equation}
h_p(s) = \frac{1}{\sigma\sqrt{2\pi}} e^{\frac{(s - d(p))^2}{2\sigma^2}} ~~~~
h'_p(s) = - \frac{s - d(p)}{\sigma^2} h_p(s) ~ .
\label{eq:proxies_statistics_slhgaussian}
\end{equation}

This compressed model stores the repartition of \addedSinceECCV{shape} inliers distances to the proxy
and makes possible estimating the \emph{diversity} of the values within each cell by counting the \emph{number of modes} in the distribution\addedSinceECCV{, appearing naturally when building the histogram.}
If it has a single mode, then all values are similar and the surface of the proxy within the cell is most likely flat.
If the distribution has two or more modes, then the values belong to different groups and the cell likely overlaps a salient area of the surface.
\addedSinceECCV{A close-up example of flat and salient proxy areas is shown as supplemental material (\autoref{S_fig:statistics_modes}), with representation of the associated smooth histograms and distribution modes.} %

\paragraph*{Color}

\addedSinceECCV{

In order to accumulate faithful color information within our coarse grid, we define color samples at higher resolution than the cells.
To do so, we take inspiration from the \emph{mesh colors} methodology \cite{yuksel2010mesh} which defines discrete color positions on faces and edges of triangle meshes.
In our framework, each cell of grid contains a fixed color sub-resolution grid which is updated by each RGB-D sample point falling into the cell.
\autoref{fig:proxies_statistics_colorpoints} details the update of color in the grid from an RGB-D sample.

\begin{figure}
\centering
    \includegraphics[width=0.8\columnwidth]{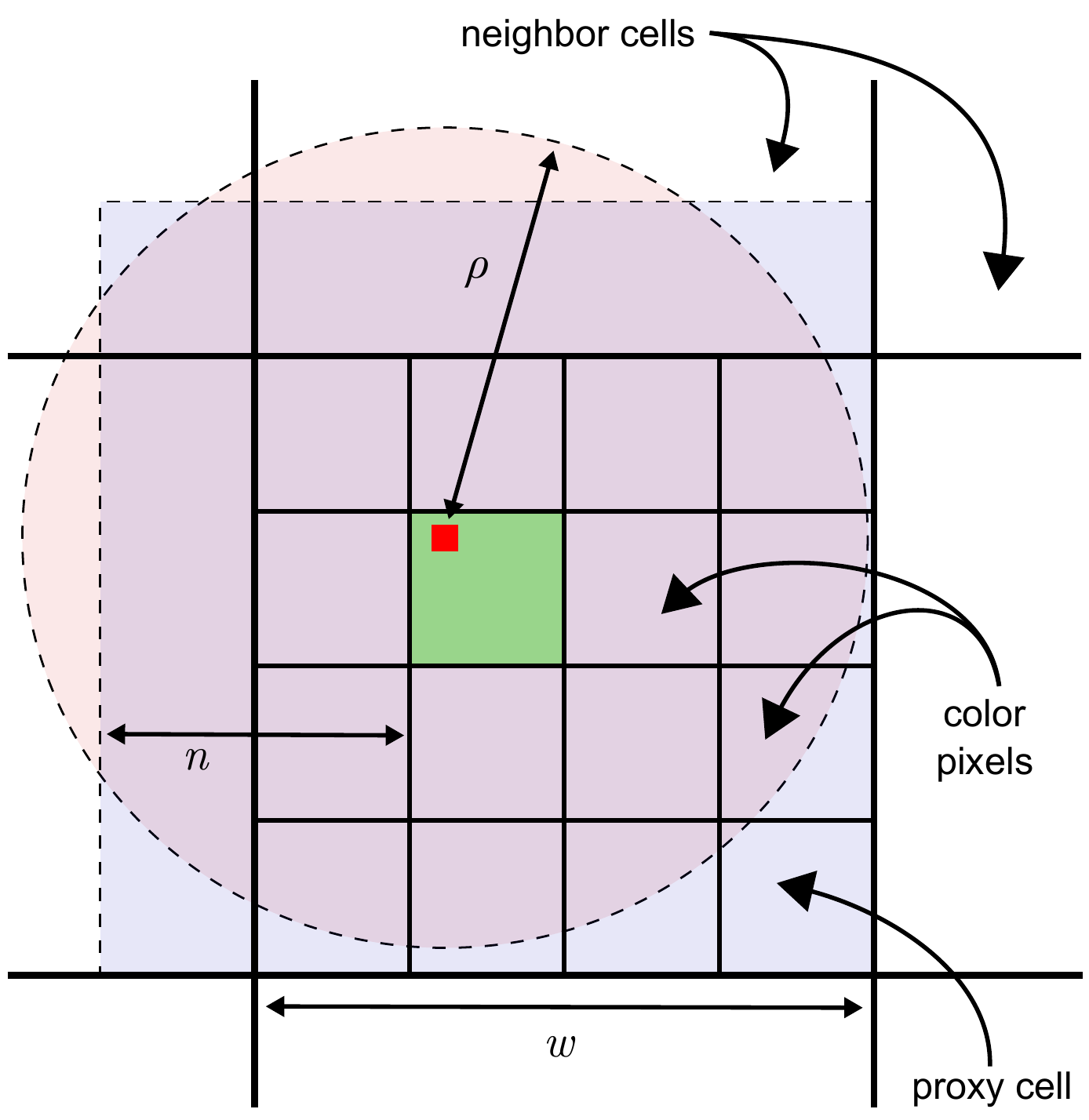}
\caption{\addedSinceECCV{Updating color points of a proxy cell from the color component of an RGB-D point sample.
An RGB-D point sample projected on the 2D proxy grid (red square) falls within a discretized color point (light green).
$w$ is the cell width in meters.
Here, $2^r$ color points are defined with $r = 2$.
Radii of influence area $\rho$ (dotted light red) and color point neighborhood $n$ (dotted light blue) are computed from the point depth (see \autoref{eq:proxies_statistics_colorneighbor}).
All color points within the computed neighborhood $n$ are updated with a weight depending on the distance to the point and the size of the influence area (see \autoref{eq:proxies_statistics_colorinfluence}).}}
\label{fig:proxies_statistics_colorpoints}
\end{figure}

For each RGB-D sample falling within the bounds of a given proxy cell, we compute a neighborhood $n$ of color positions which will be updated with this sample's color value.
In order to update as much color positions as covered by a depth point, we compute an \emph{influence radius} of the point based on its \emph{distance to the sensor}.
The topology of depth sensors implies that the further the depth point, the larger the size of the image pixel when unprojected to 3D.
Hence, the influence radius is computed as the \emph{size of this depth pixel in 3D} and discretized to get the size of the color point neighborhood at the surface of the shape.

For a 3D point sample of depth $z \in \mathbb{R}^+$, with a width of grid cells defined as $w \in \mathbb{R}^+$ and a color point resolution defined by its power of two $r \in \mathbb{N}$,
the point influence radius $\rho$ and discrete color neighborhood radius $n$ are given by

} %

\begin{equation}
\addedSinceECCV{ %
\begin{array}{l}
\rho = \frac{z}{2} \tan(\frac{fov_H}{res_H}) \in \mathbb{R}^+\\ %
n = \left\lfloor \rho \frac{2 ^ r}{w} \right\rfloor \in \mathbb{N} ~ .\\ %
\end{array}
} %
\label{eq:proxies_statistics_colorneighbor}
\end{equation}

\addedSinceECCV{

Then, for each color point within the computed neighborhood $n$, including points belonging to neighboring cells, the average color is updated with the point's color.
In order to take into account the uncertainty of information brought by depth samples far from the camera, we \emph{weight} the color contribution
with both the \emph{distance} from the color point to the projected depth point and the color \emph{neighborhood size}, which itself is computed from the point depth, as shown in \autoref{eq:proxies_statistics_colorneighbor}.
Here, weighting the color \emph{inversely} to the point depth allows further points to give a less important contribution to the color model.
For color point neighbor $(u,v) \in \llbracket -n,n \rrbracket \times \llbracket -n,n \rrbracket$,
the color contribution weight of a given depth point is defined in \autoref{eq:proxies_statistics_colorinfluence}.
In practice, we empirically define the standard deviation factor $\alpha = 3$, which gives a good balance between consistency and sharpness of the textures.

} %

\begin{equation}
\addedSinceECCV{ %
\frac{1}{1 + 2 n^2}  e^{- \frac{\| (u,v) \|^2}{2 \sigma^2}} \text{ with } \sigma = \alpha 2^r, \alpha \in \mathbb{R}^+
} %
\label{eq:proxies_statistics_colorinfluence}
\end{equation}

\section{RGB-D Stream Processing}
\label{sec:processing}

\addedSinceECCV{

Our first use of the consistent geometric proxy structure is as a support for filtering operators with the goal of:
}

\begin{itemize}
\addedSinceECCV{ %
    \item \emph{removing noise} to smooth data while keeping small details and local saliency;
    \item \emph{filling holes} due to specular elements or unseen parts;
    \item \emph{resampling} the incoming point cloud at any desired resolution at the surface of shape proxies.
}
\end{itemize}

\subsection{Filtering}
\label{sec:processing_filtering}

While projecting the sensor's data points onto their associated proxy would allow removing the acquisition noise and quantization errors due to the sensor,
this would lead to the \emph{flattening} of all \addedSinceECCV{shape} inliers.
In order to minimize the loss of details on the \addedSinceECCV{shape} surfaces while keeping a lightweight data structure,
we instead use the \addedSinceECCV{geometric} proxies as a simple \emph{collaborative filter} model.

To that end, we designed a custom filter to leverage the \emph{smoothed local histograms} stored in each cell of the proxies.
As explained in \autoref{sec:proxies_statistics_local}, the number of detected modes allows \emph{distinguishing} flat areas of the proxy surface from salient ones.
For flat cells whose distribution has a single mode, we project the depth points on the \addedSinceECCV{shape} along the direction between the camera and the point.
We offset the points of the average distance to the \addedSinceECCV{shape} only if it is above the noise threshold at the corresponding distance to the camera (see details on the noise threshold in the supplemental material, \autoref{S_sec:experiments_implementation_noise}). %
This allows \emph{smoothing} surface areas that are exactly on the \addedSinceECCV{shape} while \emph{keeping} flat areas offset from the \addedSinceECCV{shape} as they are in the scene.
For cells whose distribution has two or more modes, we do not perform any projection in order to keep the saliency \addedSinceECCV{and details} of the surface.

\autoref{eq:processing_slhfilter} details the \emph{smoothed local histograms}-based filtering of inlier $p$ to $p_f$,
belonging to cell $c$ with $m_c$ modes and an average distance to the proxy of $d_c$, and a noise threshold of $\alpha$.

\begin{equation}
p_f = 
\begin{cases}
\text{proj}(p) & \quad \text{if } m_c = 1 \wedge d_c \leq \alpha\\
\text{proj}(p) + d_c \text{ norm}(p)  & \quad \text{if } m_c = 1 \wedge d_c > \alpha\\
p & \quad \text{if } m_c \neq 1
\end{cases}
\label{eq:processing_slhfilter}
\end{equation}

\addedSinceECCV{
The function $\text{proj}(p)$ represents the projection of $p$ on the proxy along the camera direction,
which can also be seen as the intersection between the proxy shape and the camera ray passing through the pixel that generated point $p$.
The function $\text{norm}(p)$ represents the normal vector on the shape surface at the projected point location.
An illustration of the filtering process on a planar example is shown as supplemental material (\autoref{S_fig:pointfiltering}). %
}

\addedSinceECCV{In addition,} the proxy can also be used as a high level range space for \emph{cross bilateral filtering} \cite{kopf2007joint}, where inliers of different proxies will not be processed together.

\paragraph*{Temporal Flickering Removal}

Based on time-evolving data points, the proxies consolidate the stable geometry of the scene by accumulating observations \addedSinceECCV{from} multiple frames.
Averaging those observations over time removes temporal flickering, after a few frames only.

\subsection{Hole Filling}
\label{sec:processing_holes}

Missing data in depth is often due to specular and transparent surfaces such as glass or screens.
With our \addedSinceECCV{framework}, observed data is \emph{reinforced} over multiple frames from the support of stable proxies, \emph{augmenting} the current frame with samples from previous ones.
In practice, the depth data that is often seen in incoming frames creates activated cells with sufficient occupancy probability to survive within the model even when samples for these cells are missing.

This hole filling, stemming naturally from the proxy structure, is completed by two additional steps.
First, the extent of the proxies is \emph{extrapolated} to the intersection of adjacent proxies --
this is particularly useful to complete unseen areas under furniture for example.
Second, we perform a \emph{morphological closing} \cite{serra1983image} on the grid of cells, with a square structural element having a fixed side of seven cells.
This corresponds to closing holes of maximum 35cm by 35cm, which allows filling missing data due to small specular surfaces, e.g. computer screens or glass-door cabinets,
while keeping larger openings such as windows or doors.

\subsection{Resampling}
\label{sec:processing_resampling}

\addedSinceECCV{Our proxies can be used to} \emph{super-sample} RGB-D streams on the fly.
The low definition geometric component \addedSinceECCV{of raw frames} can be enriched by the higher resolution information structured in the proxies.
\addedSinceECCV{Our 3D structure can \emph{guide the process} using both its color and geometric components, whose smoothness and stability \emph{appear naturally} with the accumulation of samples.
In addition, the local nature of statistics stored in the proxy cells allows generating high resolution data while keeping geometrical details and salient areas and \emph{avoid over-smoothing}.
The sub-resolution color component of the proxies can be used to assist the sampling process to recover even more detail.}
This results in high definition RGB-D data with \emph{controllable} point density on the surface of the shapes.

\section{Experiments}
\label{sec:experiments}

\subsection{Validation on Synthetic Data}
\label{sec:experiments_synthetic}

\begin{figure*} %
\centering
    \includegraphics[width=\textwidth]{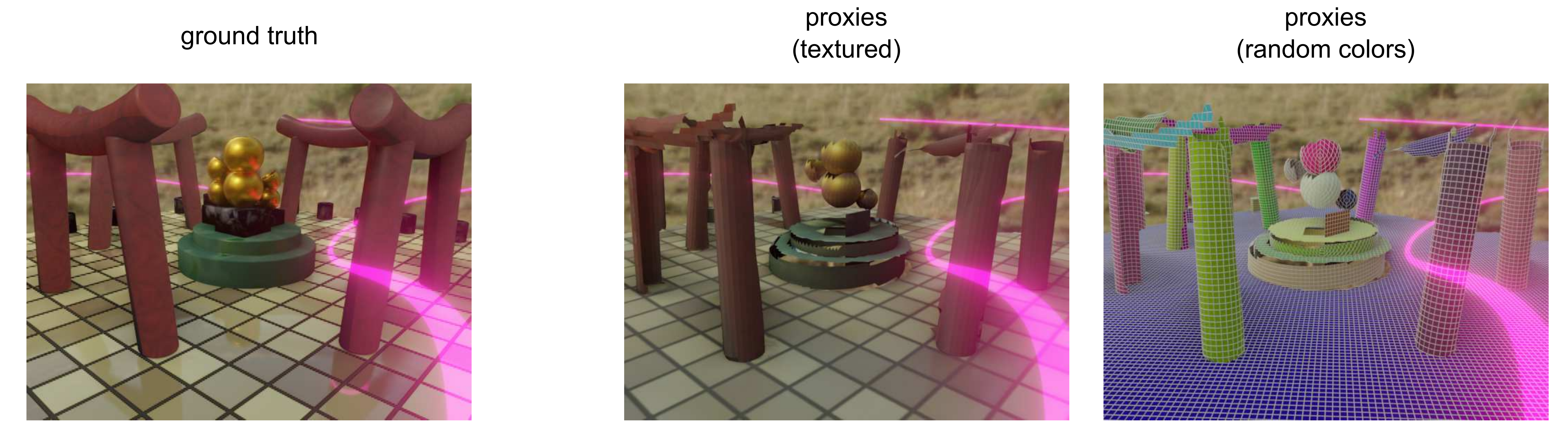}
\caption{\addedSinceECCV{Simple shapes modeling from synthetic data. %
From a set of RGB-D images (left) rendered along the pink transparent camera path, proxies are built for planar, cylindrical and spherical synthetic elements.
Both geometry and texture of dominant elements are recovered at lower resolution than the original while keeping a good visual quality (middle).
For these simple shapes, a coarse proxy grid resolution allows recovering most of the geometry (right).
More results on generated scenes are shown as supplemental material (\autoref{S_sec:synthetic}).}}
\label{fig:experiments_synthetic}
\end{figure*}

\addedSinceECCV{

We define a synthetic data processing framework,
in order to validate RGB-D data modeling with our geometric proxies while testing all kinds of shapes, even when not commonly seen in human made environments.
We use the \emph{Blender} tool\footnote{Blender: \url{https://www.blender.org/}}
to generate a synthetic RGB-D image set from known 3D objects %
of planar, cylindrical and spherical shapes, along with ground truth camera poses.
\autoref{fig:experiments_synthetic} shows the proxies generated from this artificial dataset, along with a ground truth rendering at the same position.
Results from different generated scenes are shown as supplemental material (\autoref{S_sec:synthetic}).

In particular, processing data generated from synthetic models allows comparing texture information between proxy-generated and ground truth images from 3D models.
Qualitative comparison of texture images used to generate the synthetic data and computed with the color point model of our proxies, is available as supplemental material (\autoref{S_fig:synthetictextures}). %
This figure shows that our proxies allow re-parameterization of the texture information for the revolution surfaces,
from \emph{bin packing} for the cylinder or \emph{spherical} for the sphere, into more meaningful and less distorted \emph{cylindrical} or \emph{octahedral} solutions.

}

\subsection{Building Proxies}
\label{sec:experiments_building}

\addedSinceECCV{Our proxies} are implemented through hardware and software components.
The hardware setup is made of a computer with Intel Core i7 at 3.5GHz and 10GB memory. No GPU is used.
The software setup has a client-server architecture, where the server runs within an embedded environment with low computational power
and limited memory to trigger the sensor and process the data.
The client's \emph{graphical user interface} allows controlling the processing parameters and getting a real time feedback of the stream.
A limited range of intuitive parameters allows the user to control the trade-off between quality of the output and performance of the processing.
\addedSinceECCV{In order to achieve a better quality and efficiency when building proxies, a few minor optimizations have been implemented, detailed in the supplemental material (\autoref{S_sec:experiments_implementation}).}

\subsubsection{Dataset}
\label{sec:experiments_building_dataset}

We run all of our experiments on the \emph{3DLite} \cite{huang20173dlite} dataset\footnote{3DLite: \url{http://graphics.stanford.edu/projects/3dlite/\#data}},
containing 10 scenes acquired with a \emph{Structure} sensor\footnote{Structure sensor: \url{http://structure.io}} under the form of RGB-D image sequences.
This choice was motivated by the availability of ground truth poses along with the visual data,
as well as result meshes and performance metrics provided from processing with both \emph{BundleFusion} \cite{dai2017bundle}
and \emph{3DLite} \cite{huang20173dlite}, with which we compare our method in \autoref{sec:consolidation_reconstruction}.

\subsubsection{Quantitative Results}
\label{sec:experiments_building_quantitative}

\addedSinceECCV{For the 10 scenes of the dataset, our method generates between 40 and 90 proxies per scene.
Up to 130K proxy cells per scene allow modeling the input data.}
Geometric statistics on the generated proxies are available in \autoref{tab:experiments_geomstats} for all processed scenes.
The accuracy of the proxy representation can be quantitatively assessed through the PSNR values in \autoref{tab:experiments_processing_compression}.
\addedSinceECCV{In addition}, the fast convergence of the proxy statistics is shown in \autoref{S_fig:avgincrement} of the supplemental material,
where the \addedSinceECCV{variation} over time of the average \addedSinceECCV{distance to shape falls below 0.5mm} after about 30 accumulated samples \addedSinceECCV{only}. %

\begin{table}[ht]
\setlength{\tabcolsep}{3.2pt} %
\caption{Statistics on the geometric proxies} %
\label{tab:experiments_geomstats}
\sffamily
{\centering
\begin{tabular}{ c c c c c }
    scene
    & \# proxies
    & \# cells
    & avg \# proxies / fr.
    & avg \# cells / fr.
    \\
    \hline
    apt & 51 & 62K & 3.8 & 3139 \\
    offices & 86 & 123K & 3.7& 3958 \\
    office0 & 50 & 44K & 4.0 & 2491 \\
    office1 & 56 & 53K & 4.5 & 3108 \\
    office3 & 56 & 50K & 3.9 & 2332 \\
    scene0220\_02 & 41 & 42K & 4.7 & 3369 \\
    scene0271\_01 & 32 & 33K & 4.4 & 3340 \\
    scene0294\_02 & 31 & 39K & 4.5 & 4441 \\
    scene0451\_05 & 48 & 42K & 4.7 & 3984 \\
    scene0567\_01 & 42 & 48K & 4.4 & 3354 \\
    \hline
    \\
\end{tabular}}
\emph{Total number of proxies and cells for each scene of the \emph{3DLite} \cite{huang20173dlite} dataset.
The average number of observed proxies and cells per frame (/ fr.) is given.}
\end{table}

The time required to build and update geometric proxies using our \addedSinceECCV{current} implementation is around \addedSinceECCV{150} ms for an input depth image of 320x240 pixels.
A detailed graph presenting the \addedSinceECCV{repartition of the processing time} for all steps is available as supplemental material (\autoref{S_fig:timings}). %

\subsection{Stream Processing}
\label{sec:experiments_processing}

\subsubsection{Live Enhancement}
\label{sec:experiments_processing_enhacement}

\begin{figure*} %
\centering
    \includegraphics[width=\textwidth]{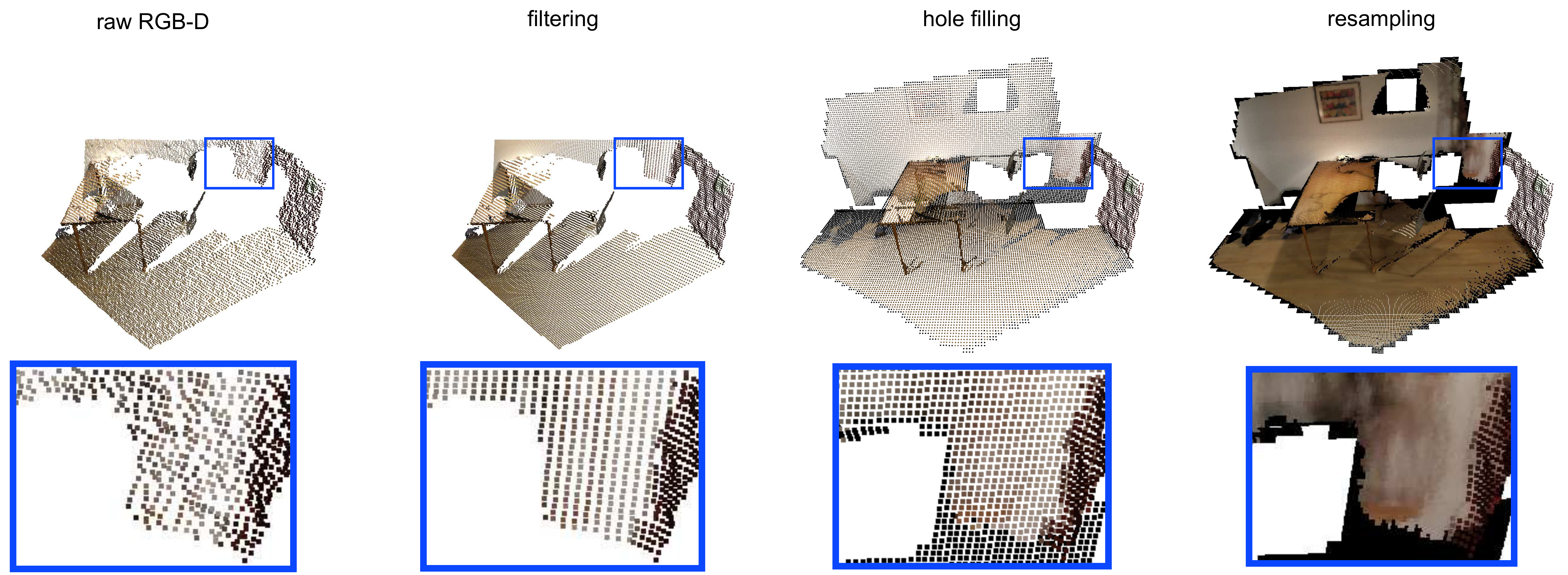}
\caption{Data improvement \addedSinceECCV{using geometric proxies.}
Raw RGB-D and proxy-improved data showing results of real time filtering, hole filling and resampling.
\addedSinceECCV{Hole filled and resampled point clouds are generated by sampling the proxy surface with respectively 2x2 and 4x4 points per cell.
Blue surrounded areas highlight a region where improvement using the proxies is significant compared to the low quality input RGB-D frames.}}
\label{fig:experiments_processing_enhacement}
\end{figure*}

\autoref{fig:experiments_processing_enhacement} shows examples of data improvement using the processing modules of our framework.
Experiments show that the proxies are particularly efficient to remove noise over walls and floors while keeping salient parts,
and help reducing holes due to unseen areas, specular areas such as lights or glass, and low confidence areas such as distant points.
Resampling the point cloud allows recovering structure if the sensor did not give enough data samples, e.g. on lateral surfaces.
\addedSinceECCV{Timings for hole filling operations are given in the supplemental material (\autoref{S_fig:timings}).} %

\subsubsection{Compression}
\label{sec:experiments_processing_compression}

Compression of the input data is achieved by using directly the proxies as a compressed, lightweight geometric substitute to the huge amount of depth data carried in the stream,
avoiding storing uncertain and highly noisy depth regions, while still being able to upsample back to high resolution depth using a bilateral upsampling.
In particular, this is convenient to broadcast captures of indoor scenes where planar regions are frequent.

Substituting the \addedSinceECCV{geometric proxies} to the RGB-D stream provides a simple yet effective lossy compression scheme for transmission,
with the practical side effect of removing many outliers.
Our efficient data structure leads to good compression ratios while keeping high \emph{peak signal-to-noise ratio (PSNR)}
and being fast for compression and decompression.
\autoref{tab:experiments_processing_compression} and \autoref{tab:experiments_processing_compressiontimings} show evaluation metrics of the compression using proxies.

The proxies \addedSinceECCV{are designed with a unified parameterized 2D structure and} are stored as simple grids of statistics with a local frame and bounding rectangle.
As such, the compressed structure itself, i.e. the proxies, can benefit from image-based compression schemes such as \emph{JPEG} \cite{wallace1992jpeg}
for offline export and storage, for which we report compression ratios and PSNR values as supplemental material (\autoref{S_tab:compressionjpeg}). %
\addedSinceECCV{The \emph{JPEG} export and load procedure for all proxies of a scene takes an average of about 40 ms.}

\begin{table}
\setlength{\tabcolsep}{6.4pt} %
\caption{Proxy compression metrics for all processed scenes} %
\label{tab:experiments_processing_compression}
\sffamily
{\centering
\begin{tabular}{ c c c c c c }
    \multirow{2}{*}{scene}
    & \multicolumn{1}{ c }{frame r.}
    & \multicolumn{2}{ c }{scene ratio}
    & \multicolumn{2}{ c }{PSNR (dB)}\\
    & Proxies & Proxies & H.264
    & Proxies & H.264\\
    \hline
    apt & 5.40 & \textbf{790.5} & 11.1 & \textbf{38.0} & 25.0 \\
    offices & 4.29 & \textbf{1173.7} & 10.5 & \textbf{38.0} & 22.1 \\
    office0 & 6.78 & \textbf{2376.6} & 14.4 & 21.4 & \textbf{30.3} \\
    office1 & 5.44 & \textbf{1828.0} & 10.5 & \textbf{41.2} & 26.9 \\
    office3 & 7.24 & \textbf{1286.2} & 12.3 & \textbf{33.8} & 29.0 \\
    scene0220\_02 & 5.02 & \textbf{814.7} & 11.1 & \textbf{36.7} & 18.3 \\
    scene0271\_01 & 5.07 & \textbf{977.8} & 10.5 & \textbf{43.8} & 18.8 \\
    scene0294\_02 & 3.82 & \textbf{1024.3} & 9.4 & \textbf{43.4} & 17.2 \\
    scene0451\_05 & 4.25 & \textbf{694.9} & 7.9  & \textbf{42.1} & 15.5 \\
    scene0567\_01 & 5.05 & \textbf{729.8} & 15.6 & \textbf{40.2} & 20.0 \\
    \hline
    \\
\end{tabular}}
\emph{Compression ratios (frame-wise and scene global) are based on the raw size of a 320x240 depth map and the size of the proxies without the outliers.
The \emph{peak signal-to-noise ratio (PSNR)} is computed using the average \emph{root mean square error (RMSE)} between raw depth points and \emph{proxy}-filtered ones.
\addedSinceECCV{These metrics do not account for the loss of outliers of detected shapes (25\% of samples in average).}
We compare our compression performance to a state-of-the-art method based on H.264 \cite{nenci2014effective} with a quality profile of 50.}
\end{table}

\begin{table}
\setlength{\tabcolsep}{9.7pt} %
\caption{Proxy compression timings for all processed scenes} %
\label{tab:experiments_processing_compressiontimings}
\sffamily
{\centering
\begin{tabular}{ c  c c  c c }
    \multirow{2}{*}{scene}
    & \multicolumn{2}{ c }{Proxies (ms)}
    & \multicolumn{2}{ c }{H.264 \cite{nenci2014effective} (ms)}\\
    & comp. & decomp.
    & comp. & decomp.\\
    \hline
    apt & 150 & \textbf{32} & \textbf{93} & 81\\
    offices & 213 & \textbf{30} & \textbf{97} & 109\\
    office0 & 161 & \textbf{19} & \textbf{112} & 146\\
    office1 & 160 & \textbf{27} & \textbf{121} & 121\\
    office3 & 150 & \textbf{32} & \textbf{111} & 116\\
    scene0220\_02 & 134 & \textbf{41} & \textbf{82} & 86\\
    scene0271\_01 & 133 & \textbf{39} & \textbf{86} & 85\\
    scene0294\_02 & 152 & \textbf{39} & \textbf{100} & 84\\
    scene0451\_05 & 133 & \textbf{41} & \textbf{111} & 80\\
    scene0567\_01 & 154 & \textbf{28} & \textbf{82} & 73\\
    \hline
    \\
\end{tabular}}
\emph{\addedSinceECCV{Timings are reported in milliseconds to process a single RGB-D frame.}
The compression time is the building of proxies averaged over all frames,
while the decompression time is the generation of a depth map by applying visibility tests to the proxies.
The \addedSinceECCV{frame-wise} compression and decompression times for the method based on H.264 \cite{nenci2014effective} \addedSinceECCV{were computed from the full frame set timing.
While our Proxies add little overhead for construction, they are more than twice as fast as H.264 for decompression, allowing live display on low end platforms.}}
\end{table}

In addition to the bandwidth saving, the compressed proxy representation enables smooth \emph{super-sampling} of the geometric data,
where the output point cloud density over proxy surfaces \emph{can be increased as desired}.
The \addedSinceECCV{geometric surface} parameterization of each proxy offers a suitable domain for point upsampling operators,
while a similar approach performed directly on the RGB-D stream is blind to the scene structure.

\section{RGB-D Data Consolidation}
\label{sec:consolidation}

\subsection{Proxy-based Reconstruction}
\label{sec:consolidation_reconstruction}

While being lightweight and fast to compute, the \addedSinceECCV{proxies} represent a superstructure modeling the dominant regular \addedSinceECCV{geometric} elements of indoor scenes.
In addition to being used to filter the input point cloud and generate an enhanced RGB-D stream as output,
proxies themselves are a way to \emph{consolidate} the RGB-D frames.
Hence, meshing the proxy cells leads to a lightened organized structure and aggregating all proxies in a global space
allows reconstructing a high quality surface model of the observed scene, generated on the fly.
\addedSinceECCV{

\paragraph*{Meshing} %

In practice, our meshing process iterates over all active cells of a given proxy to connect adjacent cells.
However, revolution surfaces such as cylinders and sphere have a periodic nature
and cells at opposite limits of the parameterized domain are actually adjacent in the Euclidean domain.
Hence, to avoid discontinuity of the proxy surface mesh, we designed a closing methodology through a range check
where cells at extremities of the 2D domain are connected to cells at the other extremities.
In particular, we can see these connections in the lower part of the octahedral sphere in \autoref{fig:proxies_parameterization} (c).
The average time required to mesh the full proxy set is below 10ms, as shown in \autoref{S_fig:timings} of the supplemental material.

}

\begin{figure*} %
\centering
\includegraphics[width=\textwidth]{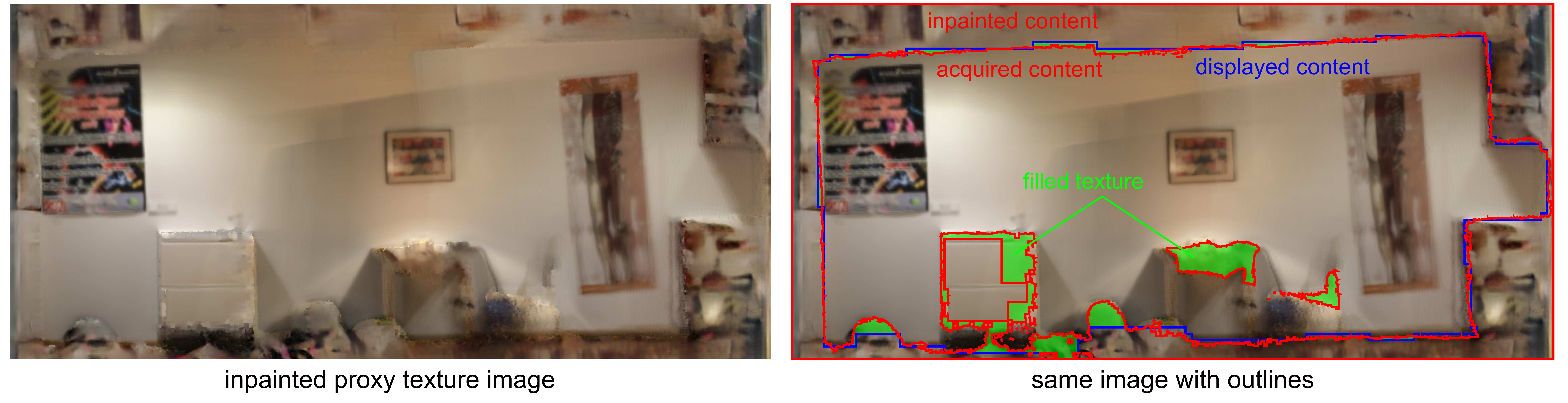}
\caption{
\addedSinceECCV{Example of an \emph{inpainted} proxy texture image from the \emph{apt} scene.
Left: Full \emph{inpainted} proxy texture image.
Right: Same image with overlays showing inpainted and displayed pixels.
The red outline shows the limit between acquired color information (inside) and inpainted content (outside).
The blue outline shows the existing proxy cells that are actually displayed in 3D.
The green overlay is the intersection of red and blue areas and corresponds to inpainted content that is actually displayed in 3D.
These green areas are geometric parts that were not observed by the camera and were activated during the hole filling step.
We can see that inpainted content at the borders of the image is not very meaningful, but gets better when surrounded by acquired data.
This figure shows that the failure of inpainting to predict good colors at border locations is not an issue for the proxies,
as most used inpainted content corresponds to filled holes surrounded by acquired colors that seem to help generating better colors.}}
\label{fig:inpainting}
\end{figure*}

\addedSinceECCV{

\paragraph*{Texture Completion}

Extrapolating grid cells in the space of the proxies, as detailed in \autoref{sec:processing_holes}, allows recovering unseen geometry but not appearance.
Hence, we apply a recent deep learning based \emph{image inpainting} method \cite{yu2018generative}
to generate meaningful pixel values at locations where the RGB-D sensor did not provide color information.
In the 2D space of the proxies, exported textures have the structure of regular images,
allowing the direct application of such off the shelf inpainting tools, as shown in \autoref{fig:inpainting}.

}

In \addedSinceECCV{the following}, we compare the performance and quality of scene reconstruction using our proxies
to state-of-the-art methods \emph{BundleFusion} \cite{dai2017bundle} and \emph{3DLite} \cite{huang20173dlite}.
\addedSinceECCV{As in the previous section, we run all of our experiments on the \emph{3DLite} dataset composed of 10 RGB-D scenes captured with a \emph{Structure} sensor (\autoref{sec:experiments_building_dataset}).}

\subsection{Qualitative Results}
\label{sec:consolidation_qualitative}

\begin{figure*} %
\centering
    \includegraphics[width=\textwidth]{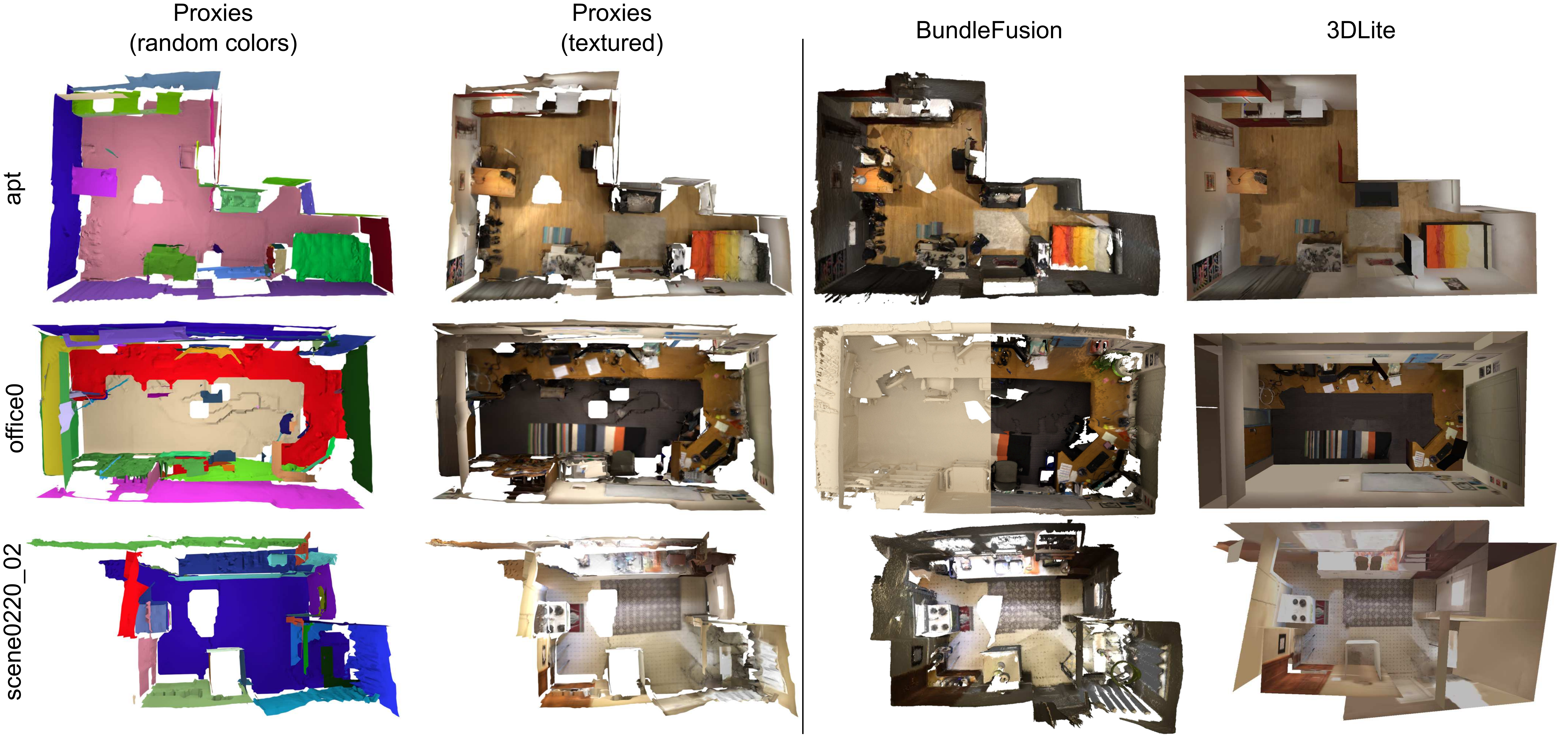}
\caption{Left: examples of reconstructed scenes of the \emph{3DLite} dataset using our geometric proxies.
\addedSinceECCV{Each proxy is shown with a different random color and textures generated from the color points stored in the proxies (see \autoref{sec:proxies_statistics_local}).}
The meshes are made of quads when four activated cells are adjacent, and triangles otherwise.
\addedSinceECCV{Right: qualitative comparison of reconstruction using our Proxies, \emph{BundleFusion} \cite{dai2017bundle} and \emph{3DLite} \cite{huang20173dlite}.
\emph{BundleFusion} meshes, while being the most accurate and containing lots of details, are a heavy model and keep some noise and outliers from the incoming data.
In addition, they lack knowledge of the scene structure and global geometry which prevents filling missing data in unobserved areas.
\emph{3DLite} meshes have geometric details smoothed out because of the strong planar regularization.
However, the textures are sharper and appearance details have a better quality.
With a \emph{good balance} between the two, proxy meshes are aware of the scene structure and composed of meaningful geometric elements.
The \emph{local} nature of the grid and its \emph{accumulated} statistics retain saliency details at the surface of the shapes.}}
\label{fig:consolidation_compare}
\end{figure*}

\autoref{fig:consolidation_compare} presents the reconstructed geometric models based on the corresponding \addedSinceECCV{proxies}.
As we can see, most large planar surfaces such as walls and floors are modeled with a \emph{single} proxy instance.
\addedSinceECCV{At scene scale, the relatively low color resolution of our proxy is sufficient to identify most elements of the scene.
The figure also compares reconstructions
with Proxies, \emph{BundleFusion} and \emph{3DLite} from a global point of view.
Qualitatively, proxy meshes are similar to \emph{BundleFusion} while offering the structural regularity of \emph{3DLite} models.
A close-up compared view is available as supplemental material (\autoref{S_fig:consolidation_closeup}).} %

\subsection{Quantitative Results}
\label{sec:consolidation_quantitative}

\autoref{tab:consolidation_reconsquanti} presents performance and quality metrics for the 10 scenes of the dataset.
The \addedSinceECCV{accuracy} of the reconstruction using proxies can be quantitatively assessed and compared to \emph{3DLite} through the values of \emph{RMSE} with \emph{BundleFusion} as a reference.
These metrics show that the lightweight and simple structure of the proxies leads to better performance both in timing and memory consumption,
while keeping a quality comparable to that of state-of-the art methods.

With their low runtime and memory needs, our proxies offer a lighter alternative to most recent reconstruction methods characterized
by \emph{volumetric} or \emph{deep learning} approaches, which have high requirements in computation costs and memory consumption.
The generic format and implementation of the proxies avoid the need for tedious platform-specific tuning
and make them well suited for embedded operation and \emph{modern mobile applications}.
In addition to the fact that our proxies are built and updated on the fly, the processing \addedSinceECCV{does \emph{not} run on a GPU} and requires far less memory than modern embedded devices offer.

\begin{table*} %
\setlength{\tabcolsep}{7.4pt} %
\caption{Quantitative comparison of proxy-based scene reconstruction} %
\label{tab:consolidation_reconsquanti}
\sffamily
{\centering
\begin{tabular}{ c  c  c c c  c c  c c c  c c }
    \multirow{2}{*}{scene}
    & \multirow{2}{*}{\# frames}
    & \multicolumn{3}{c}{processing time \addedSinceECCV{per frame (ms)}}
    & \multicolumn{2}{c}{\addedSinceECCV{max} memory consumption}
    & \multicolumn{3}{c}{model size (MB)}
    & \multicolumn{2}{c}{RMSE (m)}
    \\
    & & BF & 3DL & Proxies & BF (GB) & Proxies (MB) & BF & 3DL & Proxies & 3DL & Proxies\\
    \hline
    apt           & 2865 & --            & 6 911  & 150  & --   & 226          & 70  & \textbf{9.8} & 10.7         & \textbf{0.14} & 0.15 \\
    offices       & 8518 & --            & 4 564  & 213  & --   & 432          & 58  & \textbf{19}  & 21.8         & \textbf{0.14} & 0.18 \\
    office0       & 6159 & \textbf{26.4} & 2 104  & 161  & 21.4 & \textbf{239} & 238 & \textbf{6.3} & 8.2          & 0.14          & \textbf{0.11} \\
    office1       & 5730 & \textbf{27.7} & 1 948  & 160  & 21.1 & \textbf{241} & 251 & \textbf{7.2} & 10.3         & \textbf{0.09} & 0.15 \\
    office3       & 3820 & \textbf{27.7} & 4 241  & 150  & 16.9 & \textbf{232} & 266 & \textbf{6.2} & 9            & \textbf{0.15} & 0.19 \\
    scene0220\_02 & 2026 & --            & 4 975  & 134  & --   & 183          & 12  & 9.1          & \textbf{7.6} & 0.33          & \textbf{0.22} \\
    scene0271\_01 & 1904 & --            & 5 672  & 133  & --   & 178          & 9   & 5.8          & \textbf{5.6} & 0.11          & \textbf{0.10} \\
    scene0294\_02 & 2369 & --            & 7 598  & 152  & --   & 177          & 10  & \textbf{6.1} & 6.6          & 0.19          & \textbf{0.14} \\
    scene0451\_05 & 1719 & --            & 10 681 & 133  & --   & 189          & 12  & 9.2          & \textbf{8.3} & \textbf{0.10} & 0.13 \\
    scene0567\_01 & 2066 & --            & 5 402  & 154  & --   & 207          & 9   & \textbf{4.5} & 8.3          & \textbf{0.10} & 0.12 \\
    \hline
    \\
\end{tabular}}
\emph{The metrics are compared between \emph{BundleFusion} (BF), \emph{3DLite} (3DL) and our \emph{Proxies}.
The processing time for one frame is averaged over all frames in the scene, \addedSinceECCV{and computed from the reported full scene timing for \emph{3DLite}}.
The memory consumption is the maximum used memory during processing.
The \emph{root mean square error (RMSE)} is computed using the \emph{Metro} tool \cite{cignoni1998metro}
between the \emph{BundleFusion} mesh, taken as reference, and the \emph{3DLite} and \emph{Proxy} meshes.
\addedSinceECCV{We can see that the model generated by the proxies is similar in size and quality to the one generated by \emph{3DLite},
i.e. orders of magnitude lighter than the \emph{BundleFusion} mesh, while requiring much less computing resources.}}
\end{table*}

\section{Conclusion}
\label{sec:conclusion}

We introduced a new unified geometric framework for real-time processing of RGB-D streams,
\addedSinceECCV{based on a superstructure of proxies built from spatially and temporally consistent shape instances.}

\addedSinceECCV{By tracking geometric shapes over time and space, we define a shape-wise accumulation structure to record information brought by RGB-D samples.
The use of surface shapes instead of a volumetric structure lightens the modeling while keeping a faithful representation of most elements seen in human-made environments.
The statistics stored in the proxies maintain knowledge of the local geometry and appearance of the observed scene to keep as much detail as possible with the lowest hardware cost.}

\addedSinceECCV{We leverage the} compact, lightweight and consistent spatio-temporal support \addedSinceECCV{of these geometric proxies
within a set of} processing primitives designed to enhance \addedSinceECCV{RGB-D} data or lighten subsequent operations.
\addedSinceECCV{Our implementation} runs at interactive rates on mobile platforms and allows fast enhancement and transmission of the captured data.
Our structure can be meshed and used as a model of the observed scene, generated \emph{on the fly}.
Compared to \emph{BundleFusion} and \emph{3DLite}, reconstruction using our proxies provides a \emph{good balance} between processing time, memory consumption and approximation quality.

\paragraph*{Performance Improvements}
\label{sec:conclusion_performance}

Our current proxy model stores statistics on a uniform (yet sparse) grid,
\addedSinceECCV{which could be improved using} a sparse adaptive structure such as \emph{random access trees} \cite{lefebvre2007compressed}. %
\addedSinceECCV{
The \emph{multi-scale} nature of indoor scenes is still to be considered within our geometric proxies.
We could generate knowledge of global and local scene geometry at multiple levels, whose analysis could reveal scene information of e.g., similarities or clusters.
}
\addedSinceECCV{While we use \emph{OpenMP} to improve performance of shape processing,}
we could develop a finer parallel implementation to achieve a higher processing rate on embedded platforms. %

\paragraph*{Towards More Faithful Proxies} %
\label{sec:conclusion_faithful}

\addedSinceECCV{

Our proxies are continuously made more accurate by averaging observations in RGB-D frames, but they are highly sensitive to the quality of the camera motion estimate. %
Introducing the shape proxies into the camera motion estimation using e.g., their parameters or geometric statistics,
could give more robustness and stability to the shape tracking.

Geometric proxies model most structural components of indoor scenes, e.g. planar, however our current implementation simply ignores non regular shaped elements, such as most small objects.
In order to complete this missing information, we could define a \emph{sparse} voxel grid to \emph{locally} model the few remaining depth samples. %

Our proxy grid recovers unseen geometry at the surface of the shapes %
but sometimes fills data at locations of open doors or windows, which could be improved by modeling observed empty space or studying object semantics.

}

\paragraph*{Surface Reconstruction}
\label{sec:conclusion_regularize}

\addedSinceECCV{

We could imagine using our proxy shapes and their accumulated data as \emph{regularization priors} to be integrated into existing surface reconstruction frameworks.
We could obtain a smoother model by applying \emph{Poisson surface reconstruction} \cite{kazhdan2006poisson}
on a regularized point cloud
(preliminary results shown as supplemental material, \autoref{S_fig:smoothupsamppoisson}). %
One could also use directly the geometric proxy information as a \emph{smoothing operator} to regularize the mesh within the reconstruction process.

}

\section*{Acknowledgments}

This work is partially supported by the French National Research Agency under grant ANR 16-LCV2-0009-01 ALLEGORI and by BPI France, under grant PAPAYA.
We also wish to thank the authors of \emph{3DLite}~\cite{huang20173dlite}, \emph{BundleFusion}~\cite{dai2017bundle} and \emph{ScanNet}~\cite{dai2017scannet} for providing the dataset we use.

\bibliographystyle{IEEEtran}
\bibliography{proxyclouds}

\clearpage

\twocolumn[\begin{@twocolumnfalse}
\vskip 1cm
\begin{center}
\sffamily\Huge \emph{Supplemental Material}
\end{center}
\vskip 1cm
\end{@twocolumnfalse}]

\maketitle

\setcounter{section}{0}
\setcounter{figure}{0}
\setcounter{table}{0}
\renewcommand{\thesection}{S.\arabic{section}}
\renewcommand{\thefigure}{S.\arabic{figure}}
\renewcommand{\thetable}{S.\arabic{table}}
\renewcommand{\theHsection}{S.\thesection}
\renewcommand{\theHfigure}{S.\thefigure}
\renewcommand{\theHtable}{S.\thetable}

\section{Proxy Statistics}
\label{S_sec:proxies_statistics}

\subsection{Illustration of Smoothed Local Histograms}
\label{S_sec:proxies_statistics_slh}

\begin{figure}[ht]
\centering
    \includegraphics[width=\columnwidth]{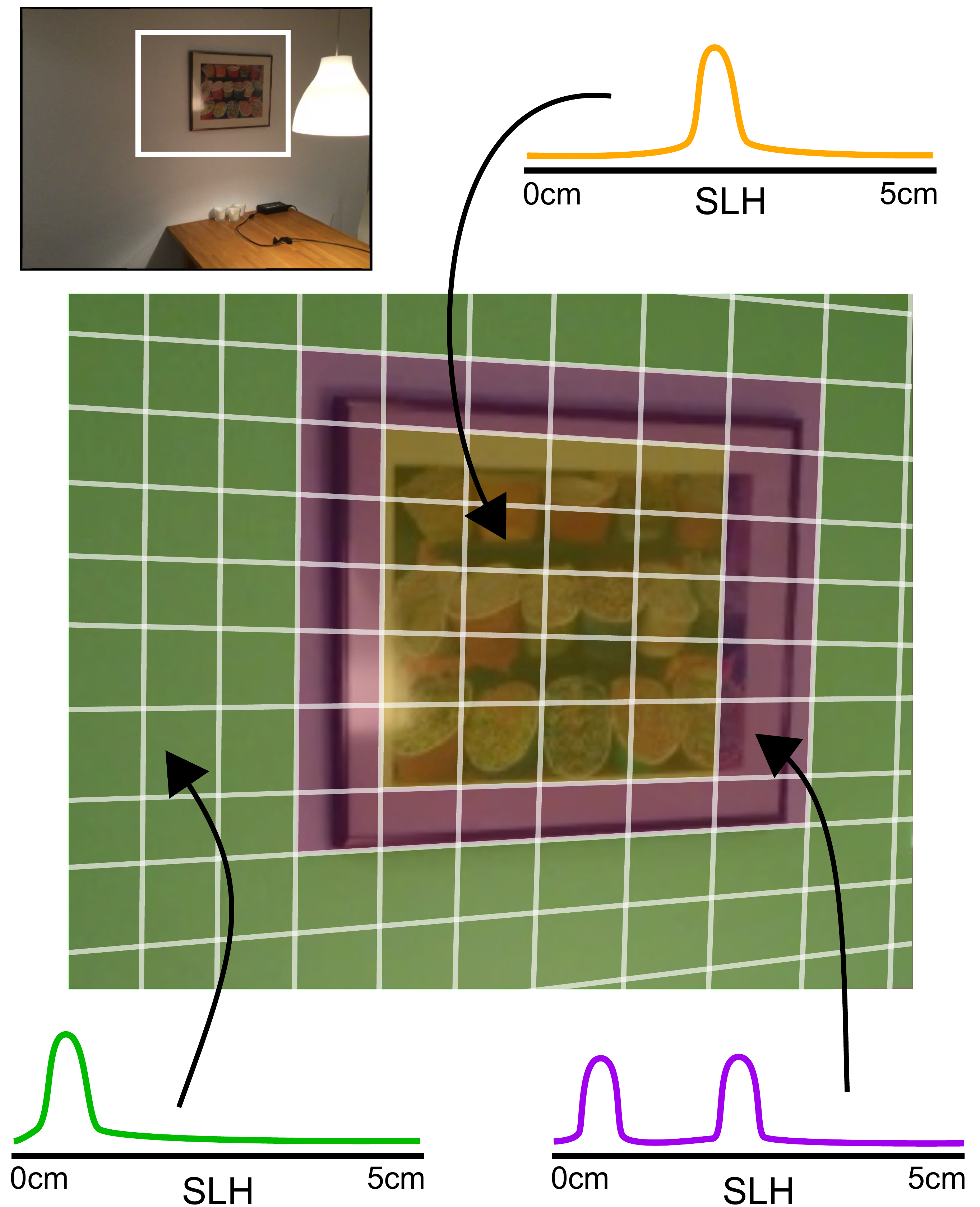}
\caption{\addedSinceECCV{Close-up example of local geometry stored in the \emph{smoothed local histogram (SLH)} of each proxy cell.
Accumulating depth samples into the histogram naturally reveals distribution modes indicating the local nature of the geometry.
Here, cells over the wall and frame have a single mode, as all samples have a similar distance to the surface (light green and orange).
Cells at the limit between wall and frame accumulate samples that can have the distance to surface of either the wall or frame, hence containing two histogram modes (light purple).}}
\label{S_fig:statistics_modes}
\end{figure}

\addedSinceECCV{\autoref{S_fig:statistics_modes} shows a close-up example of flat and salient proxy areas with representation of the associated histograms and distribution modes (\autoref{sec:proxies_statistics_local} of the paper).}

\subsection{Real Data Example}
\label{S_sec:proxies_statistics_example}

\begin{figure*} %
\centering
    \includegraphics[width=0.8\textwidth]{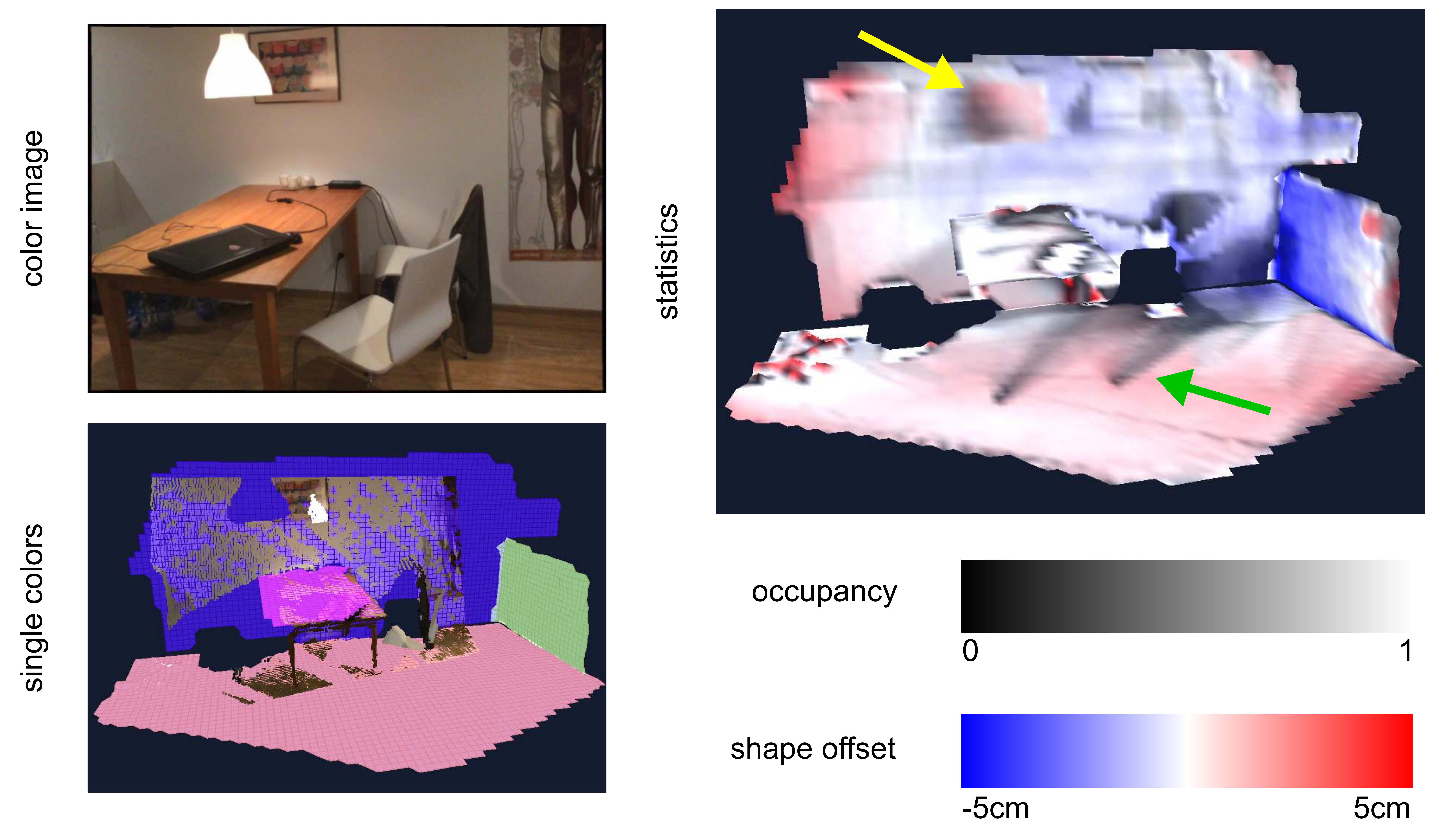}
\caption{\addedSinceECCV{Real data example of proxy statistics.
Top left: Color image provided for information in order to understand the structure of the scene.
Bottom left: Four planar proxies are represented in 3D with a single color per shape, on top of the current 3D point cloud.
Right: Proxy cells are represented with a color code showing the current values of their statistics.
Intensity represents occupancy probability, and hue -- from blue (-5cm) to red (5cm) through white (0cm) -- represents the distance to the shape.
We can notice that cells behind table legs have a lower occupancy with darker shade, as they were not observed very often (green arrow).
The geometric details due to the frame on the wall above the table are correctly recovered with a positive shape offset (yellow arrow).}}
\label{S_fig:statistics_image}
\end{figure*}

\addedSinceECCV{\autoref{S_fig:statistics_image} shows an real data example of statistics of occupancy and distance stored at the surface of our proxy shapes (\autoref{sec:proxies_statistics_local} of the paper).}

\section{Implementation Details}
\label{S_sec:experiments_implementation}

\subsection{Parallelization}
\label{S_sec:experiments_implementation_parallel}

\addedSinceECCV{
Parallelization through \emph{OpenMP} is used to process shapes faster.
When tracking shapes, as more and more previous shapes have to be checked in new frames, their score in the new frame is computed in parallel.
The update of proxies with the new shape observation is also conducted in parallel for all proxies.
However, computations such as proxy clipping or cell update within a proxy can not be parallelized because of concurrent access and inter-dependencies.
Optimization along these axes would require heavy refactoring of the current implementation.
}

\subsection{Sensor-dependent Tuning}
\label{S_sec:experiments_implementation_noise}

The axial and lateral noise introduced by the consumer depth cameras lead to artificial and erroneous data samples in the input.
Based on the sensor noise model from Nguyen et al. \cite{nguyen2012modeling}, we estimate a noise threshold at a given distance to the camera,
under which differences in geometry will not be considered as actual differences in the observed surfaces.
Hence, in order to prevent wrong geometry from being modeled within the proxies,
we modulate the distance thresholds throughout the processing by the noise estimated at the corresponding distance to the camera origin.

\subsection{Scene Orientation}
\label{S_sec:experiments_implementation_orientation}

We aim at building a model which is consistent and intuitive with relation to the structure of the observed scene.
We therefore orient the local frames of the proxies along the orientation of the scene, assuming that its structure follows the \emph{Manhattan world} assumption \cite{coughlan1999manhattan}.
\addedSinceECCV{This allows} the cells on the walls \addedSinceECCV{to be oriented along the} directions of the floor and orthogonal walls,
and the cells on the floor and ceiling \addedSinceECCV{to be oriented along the} directions of both walls.
In order to detect the structural orientation of walls and floor in the scene,
we look for nearly horizontal and vertical planes at the beginning of the processing, and use their orientations as a prior.

\section{Proxy Filtering}
\label{S_sec:filtering}

\begin{figure}
\centering
    \includegraphics[width=\columnwidth]{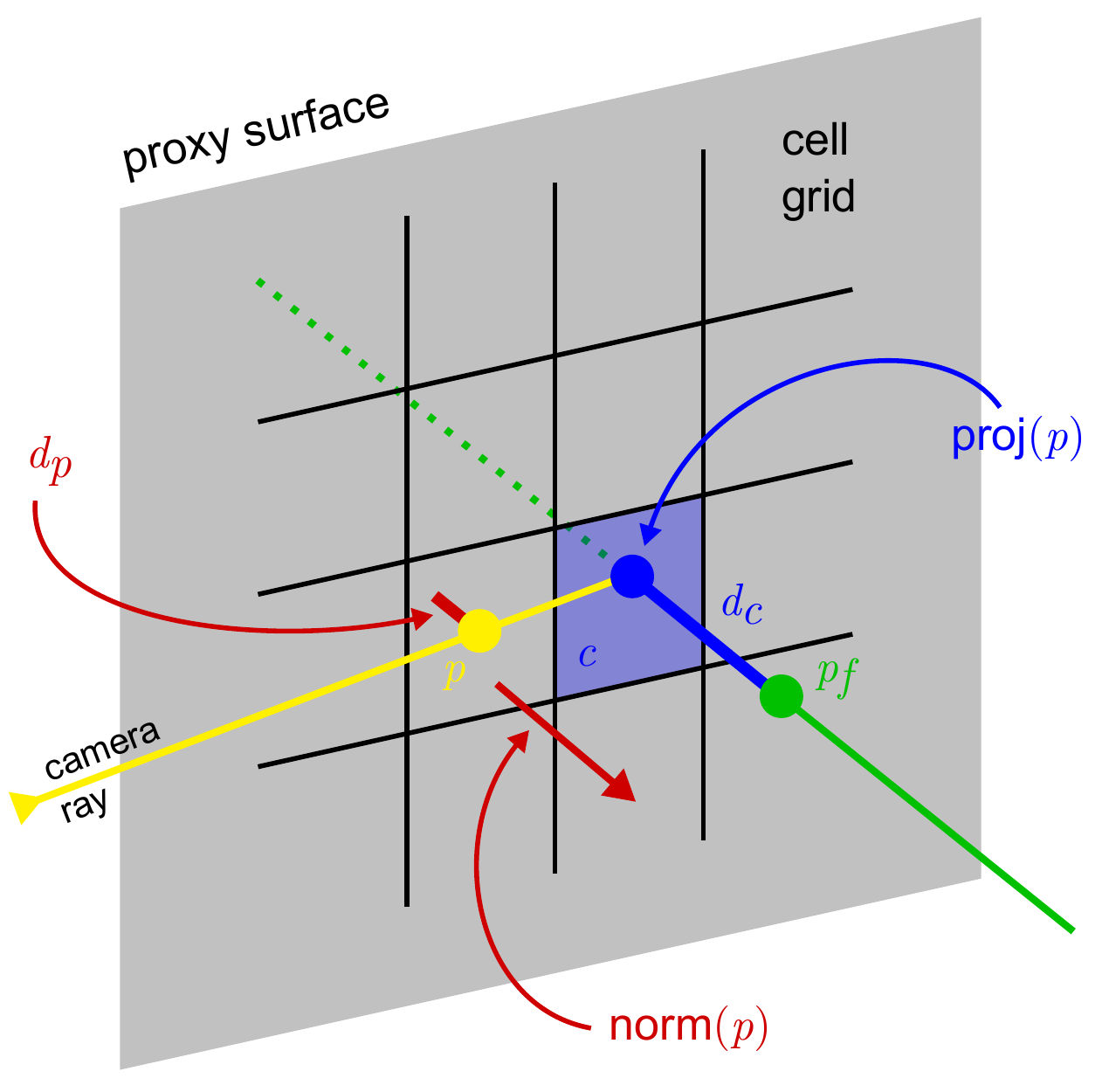}
\caption{\addedSinceECCV{Filtering 3D points using statistics from the proxy.
Example of our custom filtering on a planar proxy inlier point $p$ (yellow circle).
Here, 3D point $p$ has normal $\text{norm}(p)$ at the proxy surface (red arrow) with orthogonal distance $d_p$ (red line), used to update the average $d_c$.
Point $p$ is projected along the camera ray (yellow line) into $\text{proj}(p)$ (blue circle), which allows retrieving its cell $c$ (light blue).
In this specific example, we assume that $c$ is a cell whose \emph{smoothed local histogram} has one mode $m_c = 1$,
hence we apply the shape offset $d_c$ (blue line) along the surface normal direction (green line) to get the filtered 3D point $p_f$ (green circle).}}
\label{S_fig:pointfiltering}
\end{figure}

\addedSinceECCV{\autoref{S_fig:pointfiltering} illustrates the proxy filtering process (\autoref{sec:processing_filtering} of the paper) on a planar example
where, for a plane of normal $\vec N$ and distance to origin $l$, $\text{proj}(p) = \frac{l}{p . \vec N} p$ and $\text{norm}(p) = \vec N$.}

\section{Synthetic Data}
\label{S_sec:synthetic}

\begin{figure*} %
\centering
    \includegraphics[width=\textwidth]{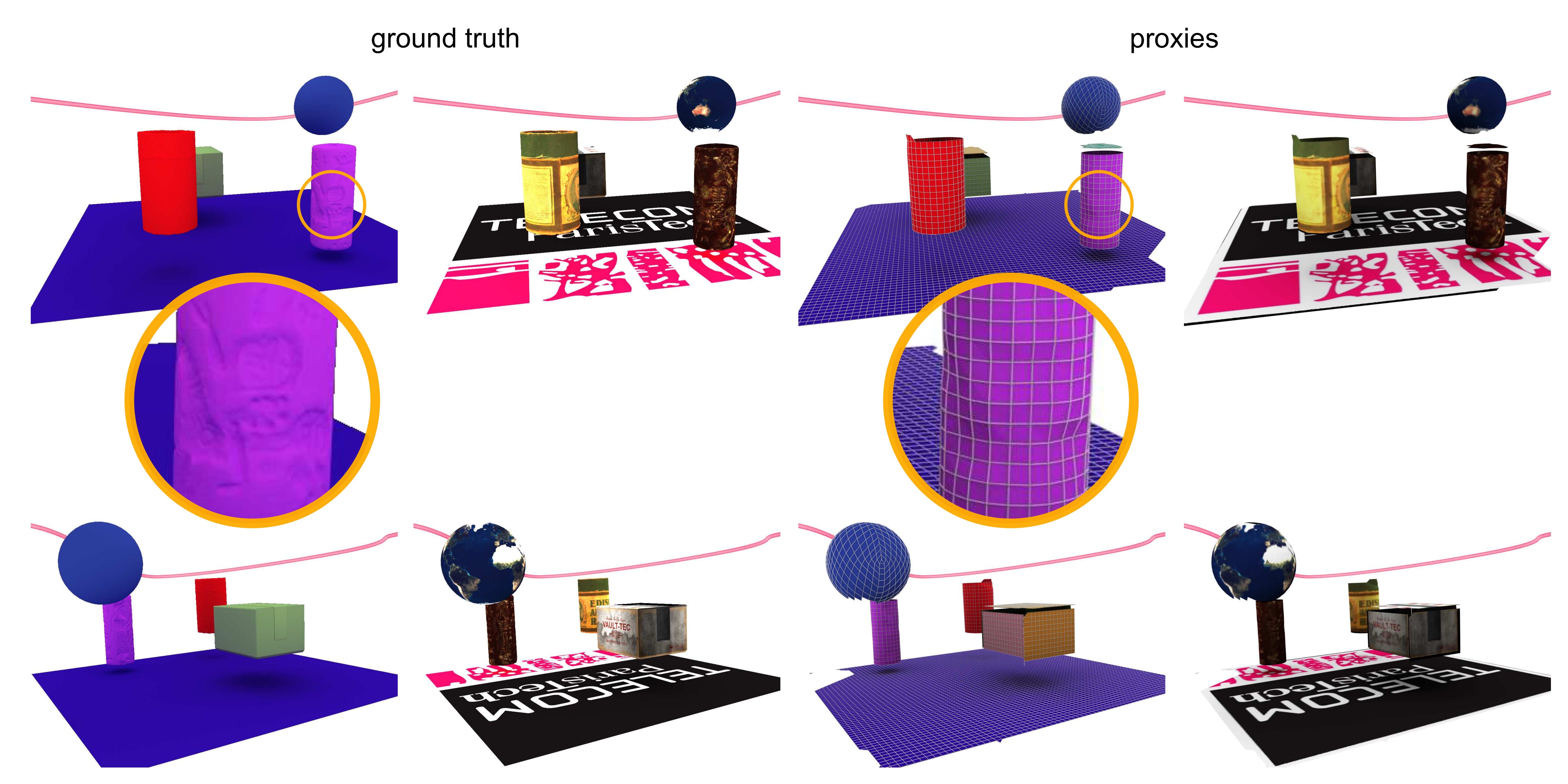}
\caption{\addedSinceECCV{Simple shapes modeling a second synthetic scene, shown at two camera locations.
The pink transparent curve represents the camera path.
Proxies are built for planar, cylindrical and spherical synthetic elements from a set of rendered RGB-D images (left).
For these simple shapes, the coarse proxy grid recovers most of the geometry and texture at lower resolution than the original (right).
The yellow circle close-ups show how proxies record local geometric information.
Although at lower resolution, we can notice slight irregularities on some proxy cells at the surface of the purple cylinder,
corresponding to fine geometric details present in the original 3D model.}}
\label{S_fig:syntheticsimplescene}
\end{figure*}

\addedSinceECCV{\autoref{S_fig:syntheticsimplescene} shows proxies generated from a second artificial dataset (details in \autoref{sec:experiments_synthetic} of the paper), along with a reference ground truth rendering at the same position.}

\begin{figure} %
\centering
    \includegraphics[width=\columnwidth]{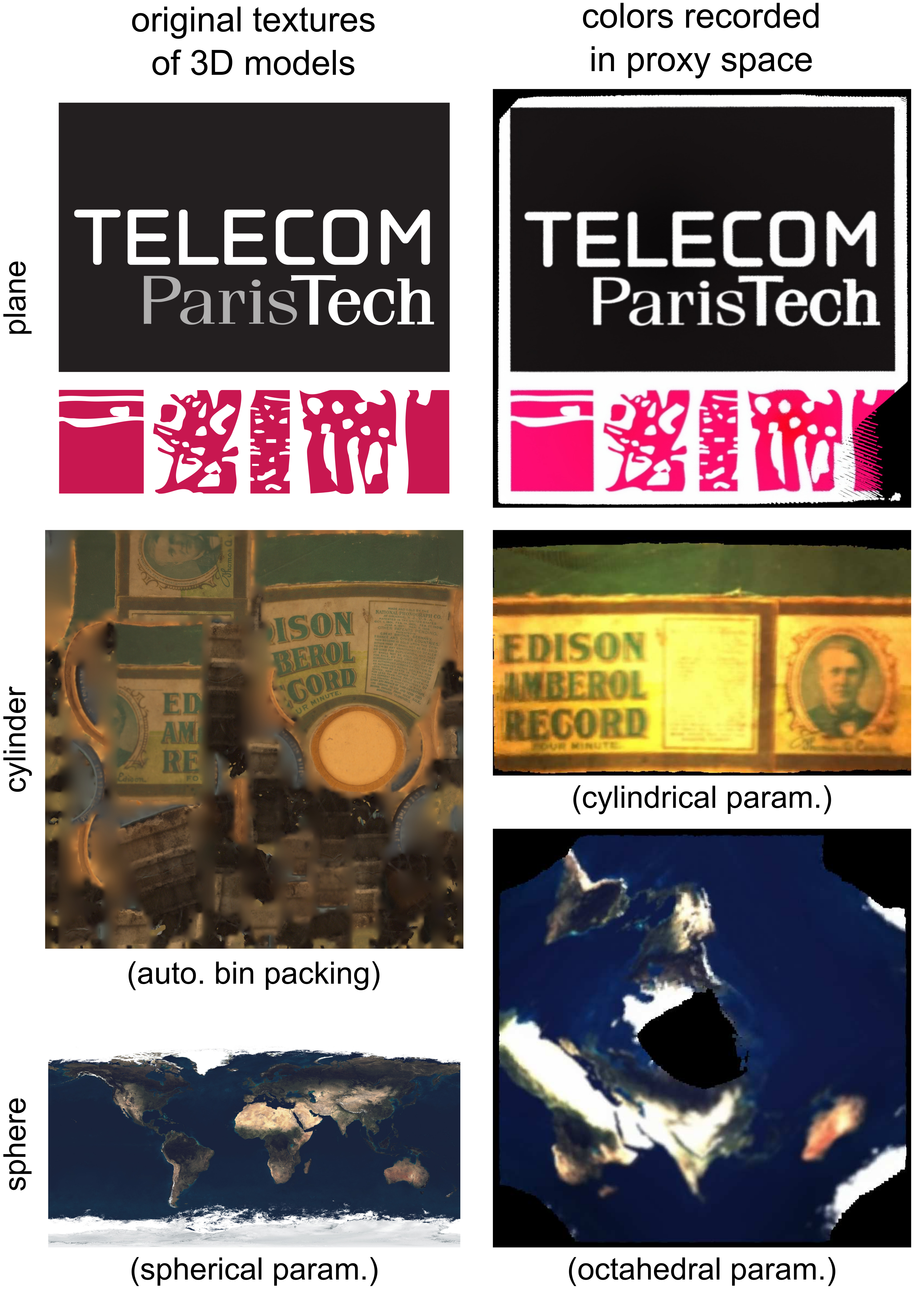}
\caption{\addedSinceECCV{Simple shape proxies modeling textures.
Left: Texture images from 3D models used to generate the RGB-D sequence in \emph{Blender}.
Right: Our local color point model embedded within the 2D grid of the proxies allows recovering textures of 3D objects.
Black areas on the proxies textures were not observed by the camera e.g., the poles of the sphere at the center and corners of the octahedron square (bottom right).
As expected, the parameterization of our proxies leads to more consistent and less distorted textures
as we can see between e.g., the octahedral and spherical sphere parameterizations (bottom).
In addition, our proxies converted the texture of the cylinder from an automatic bin packing to a more meaningful cylindrical parameterized representation (middle).}}
\label{S_fig:synthetictextures}
\end{figure}

\addedSinceECCV{\autoref{S_fig:synthetictextures} shows qualitative comparison of texture images used to generate the synthetic data and computed with the color point model of our proxies (\autoref{sec:experiments_synthetic} of the paper).}

\begin{figure*} %
\centering
    \includegraphics[width=\textwidth]{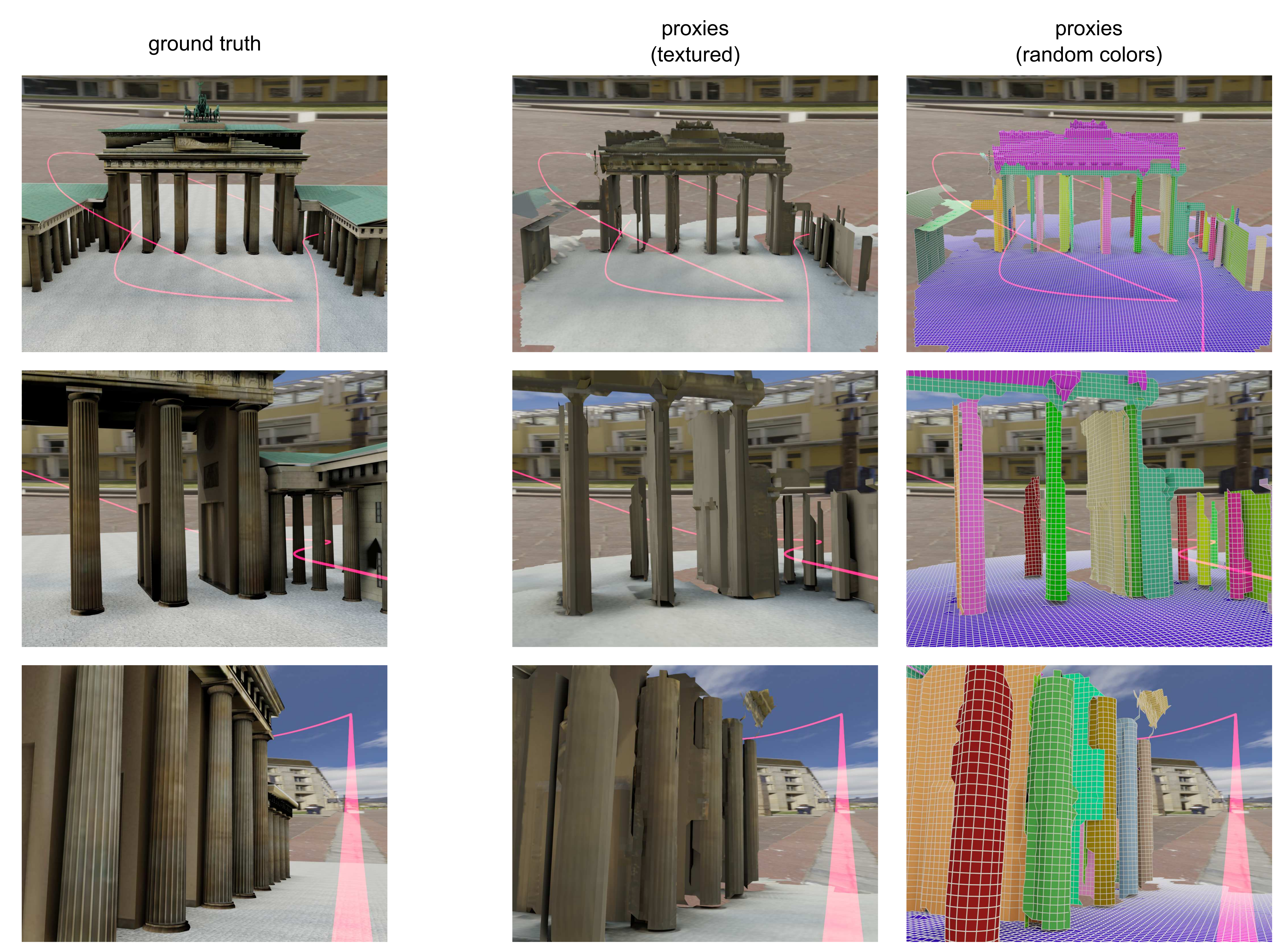}
\caption{\addedSinceECCV{Proxies generated from a scan of the \emph{Brandenburger Tor} (Berlin, Germany).
Blender was used to render RGB-D views of the scan (left) along the pink transparent camera path, three example views are shown.
Most structural elements of the landmark are recovered by planar and cylindrical proxies (right).
Data collected and processed by the \emph{CyArk} non profit organization \cite{cyark2018brandenburg}.}}
\label{S_fig:syntheticscanbrandenburg}
\end{figure*}

\addedSinceECCV{\autoref{S_fig:syntheticscanbrandenburg} shows proxies generated from a scan of the \emph{Brandenburger Tor} (Berlin, Germany)\footnote{Scan taken from \url{https://poly.google.com/view/7kolPG06JBF}} rendered in Blender, along reference ground truth renderings at the same positions.}

\section{Experiments}
\label{S_sec:experiments}

\subsection{Convergence of Proxy Statistics}
\label{S_sec:experiments_avgincrement}

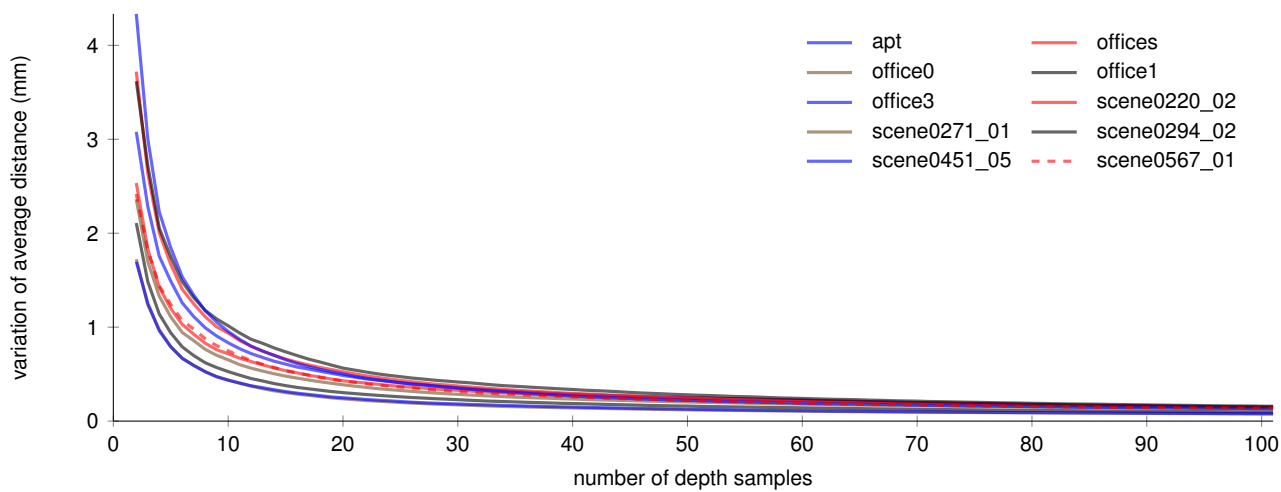
\begin{figure*} %
\centering
\begin{tikzpicture}
\begin{axis}
    [ %
    width=17cm,
    height=7cm,
    axis x line=bottom, %
    xlabel={\footnotesize\sffamily number of depth samples},
    xmin=0,
    skip coords between index={100}{200}, %
    axis y line=left, %
    ylabel={\footnotesize\sffamily \addedSinceECCV{variation of average distance} (mm)},
    ytick scale label code/.code={}, %
    ymin=0,
    axis line style={-}, %
    tick label style={font=\footnotesize\sansmath\sffamily},
    legend style={draw=none, column sep=5pt, font=\footnotesize\sffamily}, %
    legend cell align={left},
    legend columns=2
    ]

    \pgfplotstableread[row sep=\\, col sep=space]{
    cellvisitcount 2 3 4 5 6 7 8 9 10 11 12 13 14 15 16 17 18 19 20 21 22 23 24 25 26 27 28 29 30 31 32 33 34 35 36 37 38 39 40 41 42 43 44 45 46 47 48 49 50 51 52 53 54 55 56 57 58 59 60 61 62 63 64 65 66 67 68 69 70 71 72 73 74 75 76 77 78 79 80 81 82 83 84 85 86 87 88 89 90 91 92 93 94 95 96 97 98 99 100 101 102 103 104 105 106 107 108 109 110 111 112 113 114 115 116 117 118 119 120 121 122 123 124 125 126 127 128 129 130 131 132 133 134 135 136 137 138 139 140 141 142 143 144 145 146 147 148 149 150 151 152 153 154 155 156 157 158 159 160 161 162 163 164 165 166 167 168 169 170 171 172 173 174 175 176 177 178 179 180 181 182 183 184 185 186 187 188 189 190 191 192 193 194 195 196 197 198 199\\
    apt              0.00307928 0.00228562 0.00175487 0.00148999 0.00125898 0.0011132 0.000993877 0.000905784 0.000834074 0.000770991 0.000718792 0.000677191 0.000637527 0.000606422 0.000575547 0.000553805 0.000530689 0.000508649 0.000486007 0.000468527 0.000449021 0.00043538 0.000421451 0.000408211 0.000393844 0.000384746 0.000373237 0.000365313 0.000355175 0.000346915 0.000339261 0.000329842 0.000322721 0.00031562 0.000309914 0.000303585 0.000296475 0.00029112 0.000285891 0.000282233 0.000277445 0.000272169 0.000269165 0.000264525 0.000260477 0.000258522 0.00025413 0.000251623 0.000248203 0.000245538 0.000243625 0.000240699 0.000238069 0.000236462 0.000233501 0.000231608 0.000228591 0.00022622 0.000223752 0.000220587 0.000218736 0.000215394 0.000213651 0.000210378 0.000207044 0.000205405 0.000203798 0.000201095 0.000197767 0.000196126 0.000194811 0.000193084 0.000191128 0.000189264 0.000187248 0.00018485 0.000182884 0.000180697 0.000178671 0.000177088 0.000175057 0.000173237 0.000171146 0.000170222 0.000168953 0.000166448 0.000165407 0.000163398 0.000163136 0.000161257 0.000159907 0.000158742 0.00015734 0.000155813 0.000153674 0.000153011 0.000151638 0.000150429 0.00014951 0.000149053 0.00014738 0.000146104 0.000144481 0.000143271 0.000141952 0.000141231 0.000140136 0.000138969 0.000137554 0.000136675 0.000135617 0.000134826 0.000133547 0.000132525 0.000131366 0.000130348 0.000129347 0.000128142 0.000127436 0.000126277 0.000125429 0.000125027 0.000124473 0.000122873 0.000122307 0.00012191 0.000120838 0.000120219 0.000119526 0.00011829 0.000117861 0.000117098 0.000116218 0.00011529 0.000114984 0.000114602 0.000113819 0.00011304 0.000111946 0.000111063 0.000110126 0.000109712 0.00010919 0.000108079 0.00010775 0.000106935 0.000106282 0.000105485 0.000105184 0.000104702 0.000103877 0.000103164 0.000102792 0.000102347 0.000101631 0.000100953 0.000100353 9.95713e-05 9.89127e-05 9.8227e-05 9.74154e-05 9.67617e-05 9.6266e-05 9.55822e-05 9.49036e-05 9.43082e-05 9.37455e-05 9.28599e-05 9.23676e-05 9.14783e-05 9.09759e-05 9.03767e-05 8.98122e-05 8.92009e-05 8.86914e-05 8.80602e-05 8.76252e-05 8.70387e-05 8.65635e-05 8.5887e-05 8.55442e-05 8.49516e-05 8.40621e-05 8.37309e-05 8.35211e-05 8.32683e-05 8.2651e-05 8.20629e-05 8.17109e-05 8.12711e-05 8.09933e-05 8.06275e-05 8.02256e-05 7.95681e-05 7.90567e-05 7.88159e-05 7.84206e-05 7.79109e-05\\
    offices          0.00371987 0.00263904 0.00200506 0.00166894 0.00140318 0.00124873 0.00111102 0.00100302 0.000937105 0.000857123 0.000796145 0.000744738 0.000701056 0.00066079 0.00062703 0.000597965 0.000571061 0.000548879 0.000525253 0.000503533 0.000481635 0.000465054 0.000448507 0.000433235 0.000419144 0.000406549 0.00039358 0.000384421 0.00037401 0.000363458 0.000354579 0.000345956 0.00033764 0.000328981 0.000321543 0.000315211 0.000308253 0.000300729 0.000295792 0.000289904 0.000286115 0.000279697 0.000275614 0.000270136 0.000265227 0.000261637 0.000257345 0.000253876 0.000250416 0.000246625 0.000243308 0.00023913 0.000236051 0.000233556 0.000230889 0.000228088 0.000225911 0.000222697 0.000219762 0.000216655 0.000213221 0.000210723 0.000207574 0.000205092 0.000202805 0.000200404 0.00019812 0.000196251 0.000193715 0.000191559 0.000189382 0.000187444 0.00018541 0.000183076 0.000181572 0.000179928 0.000178263 0.000176828 0.000174225 0.00017254 0.000171091 0.000169015 0.000168117 0.000166613 0.000165122 0.00016336 0.000161929 0.000160454 0.000159244 0.000157978 0.000156884 0.000155346 0.000154252 0.000152686 0.000151393 0.000150082 0.000149536 0.000148362 0.000147313 0.000146004 0.000145239 0.000143854 0.000142836 0.000141884 0.000140599 0.000139227 0.000138156 0.00013745 0.000136486 0.000135526 0.000134874 0.000133756 0.000133013 0.000132046 0.0001311 0.000130678 0.000129861 0.00012906 0.00012799 0.000127388 0.000126122 0.000125378 0.000124757 0.000124132 0.00012304 0.00012241 0.000121349 0.000120598 0.000120232 0.000119351 0.000119296 0.000118616 0.000117847 0.000116731 0.000116217 0.000115551 0.000114736 0.000114006 0.00011311 0.000112552 0.00011217 0.000111295 0.000110549 0.000109873 0.000109302 0.000108951 0.000107717 0.000107247 0.000106721 0.000106234 0.000105858 0.000105172 0.000104609 0.000103991 0.000103533 0.000103058 0.000102618 0.000102152 0.000101652 0.000101168 0.000100751 0.000100234 9.98813e-05 9.93844e-05 9.86027e-05 9.81754e-05 9.78887e-05 9.7637e-05 9.73231e-05 9.68341e-05 9.6341e-05 9.60287e-05 9.56412e-05 9.50271e-05 9.46449e-05 9.4198e-05 9.39863e-05 9.35937e-05 9.30857e-05 9.26422e-05 9.22269e-05 9.2005e-05 9.16631e-05 9.14963e-05 9.08922e-05 9.0643e-05 9.0411e-05 8.99741e-05 8.97378e-05 8.95789e-05 8.91638e-05 8.89885e-05 8.8697e-05 8.84546e-05 8.81315e-05 8.78075e-05 8.75167e-05 8.70202e-05\\
    office0          0.00172246 0.00123806 0.000975119 0.000791774 0.000666692 0.000593345 0.000526106 0.000474766 0.000436634 0.000404133 0.000372256 0.000344734 0.000322379 0.000303672 0.000287319 0.000274057 0.00026035 0.000247219 0.00023713 0.000227913 0.000219108 0.000211912 0.000204707 0.000198277 0.000191709 0.000185943 0.000180709 0.000176898 0.000172305 0.000168277 0.000163783 0.000159547 0.00015622 0.000153268 0.000150608 0.000147786 0.000144852 0.000141989 0.000139778 0.000137631 0.000135938 0.000133421 0.000131448 0.000129529 0.000127097 0.000125438 0.000123837 0.000122058 0.000120002 0.000118641 0.000116894 0.000115429 0.000113724 0.000112392 0.000110915 0.000109488 0.000108487 0.000107689 0.000106455 0.000104683 0.000103851 0.000102391 0.000101094 0.00010039 9.91305e-05 9.83884e-05 9.71793e-05 9.63134e-05 9.56376e-05 9.49555e-05 9.41179e-05 9.29345e-05 9.25545e-05 9.17601e-05 9.06879e-05 9.00593e-05 8.93125e-05 8.85858e-05 8.79126e-05 8.73482e-05 8.69108e-05 8.61591e-05 8.53422e-05 8.46276e-05 8.45655e-05 8.40274e-05 8.30981e-05 8.27888e-05 8.219e-05 8.14021e-05 8.08496e-05 8.04881e-05 7.97397e-05 7.90198e-05 7.87547e-05 7.78492e-05 7.76543e-05 7.73339e-05 7.65542e-05 7.64801e-05 7.58908e-05 7.5487e-05 7.47923e-05 7.45067e-05 7.38351e-05 7.34379e-05 7.31448e-05 7.28214e-05 7.22146e-05 7.16631e-05 7.1484e-05 7.0895e-05 7.04161e-05 7.01839e-05 6.99215e-05 6.93771e-05 6.9031e-05 6.86273e-05 6.81622e-05 6.77358e-05 6.72896e-05 6.71323e-05 6.67857e-05 6.66858e-05 6.65183e-05 6.5905e-05 6.53568e-05 6.50978e-05 6.48588e-05 6.45946e-05 6.41595e-05 6.40731e-05 6.38156e-05 6.33865e-05 6.30013e-05 6.2971e-05 6.28757e-05 6.26162e-05 6.21067e-05 6.18722e-05 6.15557e-05 6.12731e-05 6.12215e-05 6.10499e-05 6.10088e-05 6.06427e-05 6.05758e-05 6.0302e-05 6.01452e-05 5.9957e-05 5.96959e-05 5.9544e-05 5.90102e-05 5.88558e-05 5.87855e-05 5.84282e-05 5.84185e-05 5.83097e-05 5.80598e-05 5.78173e-05 5.79134e-05 5.75214e-05 5.72012e-05 5.70923e-05 5.69993e-05 5.66821e-05 5.62541e-05 5.60515e-05 5.59742e-05 5.58431e-05 5.56862e-05 5.54535e-05 5.53371e-05 5.50759e-05 5.47802e-05 5.46129e-05 5.45061e-05 5.41571e-05 5.40601e-05 5.39201e-05 5.37708e-05 5.36898e-05 5.34536e-05 5.31446e-05 5.3103e-05 5.28109e-05 5.27916e-05 5.25372e-05 5.2403e-05 5.21999e-05 5.18656e-05 5.16565e-05 5.17101e-05 5.16409e-05 5.14875e-05 5.10229e-05 5.10277e-05 5.08635e-05\\
    office1          0.00210835 0.00148535 0.00114003 0.000939834 0.000790441 0.000697905 0.000621309 0.000570152 0.000526973 0.000488892 0.000452858 0.000427202 0.000399451 0.000377895 0.000358477 0.000343643 0.000329163 0.000314981 0.000303135 0.000293114 0.000283249 0.000274441 0.000266206 0.000258586 0.000251243 0.000245195 0.000238622 0.000233611 0.000228031 0.000222125 0.0002164 0.000211916 0.000208374 0.000204198 0.000200272 0.000196841 0.000193204 0.000189327 0.000187188 0.000184017 0.000180857 0.000177622 0.000174883 0.000172025 0.00016902 0.000166011 0.000163851 0.000161695 0.000160125 0.000157058 0.00015516 0.000152885 0.000151244 0.000149516 0.000147795 0.000146105 0.000144502 0.000142919 0.000141356 0.000139518 0.000137929 0.000136745 0.000134518 0.00013285 0.000131293 0.000129932 0.000128569 0.000127141 0.000125812 0.000124342 0.000123147 0.000122198 0.000120968 0.000119949 0.000118649 0.000117151 0.000116252 0.000115313 0.000114321 0.000113397 0.000112507 0.000111306 0.000110271 0.000109128 0.000108183 0.000107615 0.000106736 0.000106064 0.000105414 0.000104311 0.000103973 0.000102609 0.000101985 0.000101197 0.00010044 9.96078e-05 9.88147e-05 9.83753e-05 9.77249e-05 9.73208e-05 9.67054e-05 9.58124e-05 9.52457e-05 9.50557e-05 9.44785e-05 9.4123e-05 9.34354e-05 9.3043e-05 9.252e-05 9.1969e-05 9.151e-05 9.13045e-05 9.05266e-05 9.00767e-05 8.97119e-05 8.92888e-05 8.91201e-05 8.83891e-05 8.77888e-05 8.72053e-05 8.67901e-05 8.64742e-05 8.56921e-05 8.55487e-05 8.48377e-05 8.45947e-05 8.41308e-05 8.40194e-05 8.38102e-05 8.30719e-05 8.26863e-05 8.19637e-05 8.18044e-05 8.14436e-05 8.12917e-05 8.09091e-05 8.0679e-05 8.02987e-05 8.01792e-05 7.96849e-05 7.95287e-05 7.92601e-05 7.90877e-05 7.85509e-05 7.81837e-05 7.77955e-05 7.76528e-05 7.74006e-05 7.70221e-05 7.66711e-05 7.62888e-05 7.60801e-05 7.57274e-05 7.53418e-05 7.48412e-05 7.45747e-05 7.44174e-05 7.39507e-05 7.3392e-05 7.31708e-05 7.30521e-05 7.25018e-05 7.24343e-05 7.19948e-05 7.17291e-05 7.12793e-05 7.1011e-05 7.06406e-05 7.05011e-05 7.04188e-05 7.00265e-05 6.97389e-05 6.9874e-05 6.9864e-05 6.93806e-05 6.92216e-05 6.88739e-05 6.86345e-05 6.82187e-05 6.80712e-05 6.79908e-05 6.7693e-05 6.74882e-05 6.72921e-05 6.70416e-05 6.67407e-05 6.64441e-05 6.62466e-05 6.61729e-05 6.59074e-05 6.57543e-05 6.52971e-05 6.53083e-05 6.5169e-05 6.50252e-05 6.48817e-05 6.45663e-05 6.43599e-05\\
    office3          0.00169641 0.00125529 0.000960222 0.000791753 0.000669604 0.000592375 0.000525965 0.000471164 0.000435225 0.000404968 0.000376212 0.000355761 0.000333863 0.000314224 0.000297555 0.0002845 0.000270846 0.000257014 0.000248325 0.000240414 0.000230925 0.000222906 0.000216619 0.000208497 0.000201557 0.000195676 0.000190616 0.000186281 0.000182794 0.000178121 0.000174624 0.000171299 0.00016762 0.000164381 0.000160889 0.000157903 0.000155 0.000152161 0.000149999 0.000146658 0.000144476 0.000142094 0.000139271 0.000136803 0.000134877 0.000132649 0.000130491 0.000129 0.000126298 0.000124689 0.000122674 0.000121129 0.000119324 0.00011723 0.000115848 0.000114047 0.000112362 0.000111313 0.000109947 0.00010818 0.000106871 0.000105699 0.00010475 0.000103782 0.000102488 0.00010104 0.000100107 9.893e-05 9.7978e-05 9.71928e-05 9.64506e-05 9.57466e-05 9.48277e-05 9.40285e-05 9.29523e-05 9.22196e-05 9.18428e-05 9.07589e-05 9.04086e-05 8.96276e-05 8.87366e-05 8.81274e-05 8.79416e-05 8.70451e-05 8.6585e-05 8.59696e-05 8.54371e-05 8.4756e-05 8.42232e-05 8.35519e-05 8.29081e-05 8.23488e-05 8.20058e-05 8.17719e-05 8.10839e-05 8.04494e-05 7.9846e-05 7.91174e-05 7.86603e-05 7.80373e-05 7.77868e-05 7.74157e-05 7.69089e-05 7.63666e-05 7.5829e-05 7.53324e-05 7.47857e-05 7.41673e-05 7.38056e-05 7.36345e-05 7.28867e-05 7.23555e-05 7.20509e-05 7.15615e-05 7.11259e-05 7.08902e-05 7.07627e-05 7.00433e-05 6.97662e-05 6.9607e-05 6.9211e-05 6.89875e-05 6.8475e-05 6.8272e-05 6.76519e-05 6.73252e-05 6.70793e-05 6.68758e-05 6.65737e-05 6.64214e-05 6.58124e-05 6.54554e-05 6.50896e-05 6.509e-05 6.46152e-05 6.43246e-05 6.39774e-05 6.39748e-05 6.3596e-05 6.35543e-05 6.31222e-05 6.2966e-05 6.26358e-05 6.2442e-05 6.2057e-05 6.18368e-05 6.16266e-05 6.14623e-05 6.12798e-05 6.09303e-05 6.07795e-05 6.0366e-05 6.01797e-05 6.00444e-05 5.97498e-05 5.95913e-05 5.9309e-05 5.89824e-05 5.85452e-05 5.85212e-05 5.83089e-05 5.78304e-05 5.77228e-05 5.75013e-05 5.74256e-05 5.7065e-05 5.67619e-05 5.65669e-05 5.66129e-05 5.63951e-05 5.61328e-05 5.61029e-05 5.58349e-05 5.57961e-05 5.53883e-05 5.50016e-05 5.48723e-05 5.45823e-05 5.45142e-05 5.44175e-05 5.40988e-05 5.38202e-05 5.36851e-05 5.34265e-05 5.34199e-05 5.30635e-05 5.27028e-05 5.25961e-05 5.24138e-05 5.20646e-05 5.18624e-05 5.17929e-05 5.14901e-05 5.13561e-05 5.10759e-05 5.12039e-05 5.09834e-05 5.073e-05\\
    scene0220_02    0.00253414 0.00183226 0.00142434 0.00119879 0.00102926 0.000925001 0.000829561 0.000759354 0.000716711 0.000669778 0.000633303 0.000599427 0.000565352 0.000536811 0.000513613 0.000488577 0.000468462 0.000446948 0.000428965 0.000416611 0.000405463 0.000392475 0.00038004 0.000370315 0.000360031 0.000350782 0.000341892 0.000334154 0.000327429 0.000320241 0.000312131 0.000304339 0.000299396 0.000292396 0.000287684 0.00028159 0.000276127 0.000272612 0.000267793 0.000264266 0.000261489 0.000257898 0.000253505 0.000251251 0.000248044 0.000245543 0.000241921 0.000239482 0.000236334 0.000233817 0.000231248 0.000228177 0.000226077 0.000223656 0.000221293 0.000218355 0.000216308 0.000214825 0.000212533 0.000210053 0.000208058 0.000206391 0.000204465 0.000202282 0.000201464 0.000199335 0.000197369 0.000196071 0.00019383 0.000191523 0.000189792 0.000188172 0.00018696 0.000185234 0.000183716 0.00018243 0.000180783 0.000178489 0.000176233 0.000174421 0.000172104 0.000170564 0.000168682 0.00016733 0.00016589 0.000164074 0.000162517 0.000160648 0.00015958 0.000158468 0.000156707 0.00015569 0.000154388 0.000152987 0.000151823 0.000150969 0.000150109 0.000148954 0.000147694 0.000146822 0.00014615 0.000144815 0.000143682 0.000142185 0.000141003 0.000139505 0.000138628 0.000138142 0.000136672 0.00013513 0.000134323 0.000133801 0.000132588 0.000131743 0.000130982 0.000130129 0.000129134 0.000128445 0.00012816 0.000126745 0.000125922 0.00012485 0.000124041 0.00012309 0.000122181 0.000121707 0.000121075 0.000120121 0.000119227 0.000119 0.000118019 0.000117634 0.000116903 0.000116306 0.000115848 0.000115342 0.000114881 0.000114414 0.000113725 0.000113308 0.000112699 0.000112176 0.000112083 0.000111075 0.000109907 0.000109615 0.000109224 0.000109254 0.000108401 0.000107778 0.000107261 0.000106504 0.000106334 0.000106032 0.000105494 0.000105021 0.000104549 0.000103866 0.00010343 0.000102996 0.000102915 0.000102337 0.000101871 0.000101125 0.000100409 9.99511e-05 9.95487e-05 9.93008e-05 9.88218e-05 9.81781e-05 9.73967e-05 9.68647e-05 9.63884e-05 9.59302e-05 9.57082e-05 9.55343e-05 9.46073e-05 9.41325e-05 9.34938e-05 9.33907e-05 9.28143e-05 9.24959e-05 9.19454e-05 9.14249e-05 9.08325e-05 9.02612e-05 8.97461e-05 8.95629e-05 8.90949e-05 8.86326e-05 8.8323e-05 8.75243e-05 8.72621e-05 8.66962e-05 8.64269e-05 8.6127e-05 8.54305e-05 8.52965e-05\\
    scene0271_01    0.00236054 0.00169971 0.00132964 0.00110975 0.000943614 0.000856487 0.000763998 0.00070011 0.000652591 0.000600762 0.000563623 0.000532604 0.000504247 0.00047864 0.000457788 0.000437286 0.000419037 0.000402313 0.000386729 0.000372417 0.000359027 0.000345867 0.000335282 0.000326598 0.000316247 0.000307407 0.000299195 0.00029094 0.000285306 0.000277997 0.000271832 0.000266284 0.000260568 0.000254716 0.000251041 0.000246313 0.000243235 0.000237868 0.000234629 0.00023064 0.000226312 0.000223405 0.000220292 0.000217124 0.000214063 0.000211413 0.00020917 0.000205649 0.000203519 0.000201171 0.000199012 0.000196189 0.000193902 0.000190479 0.000188304 0.000185303 0.000183094 0.000181418 0.000179619 0.000177229 0.00017558 0.000173625 0.000171502 0.000169531 0.000167596 0.000166281 0.000164914 0.000163049 0.000162516 0.000160943 0.000159857 0.000158332 0.0001569 0.000155476 0.000154218 0.000153374 0.000151896 0.000150825 0.000149113 0.000148329 0.000147449 0.000146594 0.000145361 0.000143855 0.00014307 0.000141728 0.000140668 0.000139547 0.000138373 0.00013744 0.000137087 0.000136031 0.000134979 0.000133537 0.000132437 0.000131674 0.000130591 0.000129706 0.000128885 0.000128368 0.00012702 0.000126381 0.000125424 0.000124696 0.0001234 0.000122576 0.000121934 0.000121195 0.000120325 0.000119552 0.000118773 0.00011792 0.000117934 0.000116999 0.000116267 0.000115576 0.000115094 0.000114545 0.000114006 0.000113218 0.000112533 0.000112108 0.000111233 0.00011082 0.000110463 0.000109931 0.000109108 0.000108693 0.000108293 0.000107638 0.000106894 0.00010647 0.000105507 0.000105216 0.00010516 0.000104624 0.000103959 0.000103866 0.000103099 0.000102346 0.0001018 0.000101405 0.000100886 0.000100751 0.000100532 0.000100247 9.98511e-05 9.91223e-05 9.86461e-05 9.80897e-05 9.79552e-05 9.71162e-05 9.66512e-05 9.67119e-05 9.62065e-05 9.57481e-05 9.53581e-05 9.48533e-05 9.45159e-05 9.41128e-05 9.39619e-05 9.35551e-05 9.30573e-05 9.27461e-05 9.2467e-05 9.19221e-05 9.14115e-05 9.09433e-05 9.08186e-05 9.0791e-05 9.02524e-05 8.99002e-05 8.96913e-05 8.92249e-05 8.87714e-05 8.82828e-05 8.78381e-05 8.75721e-05 8.74787e-05 8.72805e-05 8.70096e-05 8.6711e-05 8.62227e-05 8.59127e-05 8.56547e-05 8.49112e-05 8.50204e-05 8.46031e-05 8.41204e-05 8.34629e-05 8.33372e-05 8.26494e-05 8.2609e-05 8.23075e-05 8.19528e-05 8.17575e-05 8.15249e-05 8.1267e-05\\
    scene0294_02    0.00361874 0.00272091 0.00205422 0.00175045 0.00148535 0.00131505 0.00117936 0.00108709 0.00101475 0.000937999 0.000870527 0.000828244 0.000778765 0.000737621 0.000697997 0.000659714 0.000629308 0.00059598 0.000564181 0.000543082 0.000522673 0.000503197 0.000487032 0.000473126 0.000458326 0.000447476 0.00043538 0.000426092 0.00041589 0.000408098 0.000397024 0.00038632 0.000377192 0.000370711 0.000363615 0.000356425 0.000349725 0.000342398 0.00033668 0.000329071 0.000324275 0.000318933 0.000312786 0.000305439 0.000299373 0.00029469 0.000289598 0.00028414 0.000279965 0.000275105 0.000272155 0.00026755 0.000262252 0.000259754 0.000256666 0.000251428 0.000247278 0.000244231 0.000241089 0.000237831 0.00023433 0.000231637 0.000229443 0.000226978 0.000223068 0.000220627 0.000216679 0.000214684 0.000212841 0.000208862 0.000206991 0.000204714 0.000202351 0.000200263 0.000198228 0.000195372 0.000193685 0.000192144 0.000189526 0.000187795 0.000185444 0.00018326 0.000181965 0.000180756 0.000179732 0.000177328 0.000175906 0.000173991 0.000172553 0.000170733 0.000169275 0.0001675 0.000167105 0.000164934 0.000163749 0.000162047 0.000161099 0.000160661 0.000158616 0.000157663 0.000156265 0.000155534 0.000154436 0.000153813 0.000152521 0.000151341 0.000149468 0.000148094 0.000147212 0.000146696 0.000145534 0.000145284 0.000144357 0.000143138 0.000141999 0.000141662 0.000140975 0.000140574 0.000139177 0.000139241 0.000138374 0.000137214 0.000136913 0.000135639 0.000135458 0.00013502 0.000134472 0.000133657 0.000132742 0.000132138 0.000131271 0.000131136 0.00013045 0.000130017 0.000130049 0.000128929 0.000128557 0.000127899 0.000127143 0.00012641 0.000125801 0.000125249 0.000124784 0.000124284 0.000123987 0.000123417 0.000122549 0.000122548 0.000121914 0.000121878 0.000121281 0.000120469 0.000119857 0.000119348 0.00011931 0.000118482 0.000118107 0.000117895 0.00011666 0.000115737 0.000115943 0.000115282 0.000115047 0.000114507 0.000113724 0.000112923 0.000112245 0.000111994 0.000111721 0.000111143 0.000111111 0.000110392 0.000110324 0.00010988 0.00010941 0.000109035 0.00010885 0.000108604 0.000108079 0.000107887 0.000107253 0.00010664 0.000106524 0.000106165 0.000105524 0.000105255 0.000104838 0.000104358 0.000103896 0.000103202 0.000103024 0.000102771 0.000102464 0.000101884 0.000101595 0.000101357 0.000101009 0.000100662\\
    scene0451_05    0.00433523 0.00300661 0.00222739 0.00184584 0.00153614 0.0013408 0.00117387 0.00105299 0.00095491 0.000870363 0.000797626 0.000745826 0.000699563 0.000653707 0.000613061 0.000580275 0.000551113 0.000525113 0.000502535 0.000476808 0.000457564 0.000439718 0.000423976 0.00040854 0.000396013 0.000386561 0.000373942 0.000361464 0.000350309 0.000338842 0.000329568 0.000319845 0.000311695 0.000305019 0.000295969 0.000287729 0.000281589 0.000274176 0.000266299 0.000260136 0.000254887 0.000249099 0.000245907 0.000240768 0.000235454 0.000231829 0.000227498 0.00022338 0.000220502 0.000217994 0.000213107 0.000209136 0.000206693 0.000205227 0.000201903 0.000198964 0.000195711 0.000193493 0.000190161 0.000187446 0.000186189 0.000184327 0.000181593 0.00018049 0.000178968 0.000177159 0.000174465 0.000171997 0.000169467 0.000168028 0.000166039 0.000164381 0.000163562 0.000161613 0.000160012 0.000159075 0.000157412 0.000155514 0.000154538 0.00015298 0.000151737 0.000150133 0.000148333 0.000147086 0.000146398 0.000145346 0.000143564 0.000142762 0.00014251 0.000141911 0.000140876 0.000140266 0.000138239 0.000137397 0.000136363 0.000134827 0.000134295 0.000132789 0.000131844 0.000131771 0.000129966 0.000128696 0.000128455 0.000128593 0.000127443 0.000126773 0.000125678 0.000124829 0.00012459 0.000123602 0.000122784 0.000122802 0.000122203 0.000122042 0.00012168 0.000121228 0.000120673 0.000119043 0.000118686 0.000117805 0.000117016 0.00011618 0.000115533 0.000115434 0.000114484 0.000113978 0.000112986 0.000111892 0.000111395 0.000111039 0.000109869 0.000109014 0.000108332 0.000107272 0.000107096 0.000106698 0.000105544 0.000105264 0.000104616 0.000104386 0.000103557 0.000102996 0.000102217 0.000101713 0.000100808 0.000100369 9.952e-05 9.97321e-05 9.9037e-05 9.8506e-05 9.84225e-05 9.82221e-05 9.79854e-05 9.77727e-05 9.73485e-05 9.68188e-05 9.66065e-05 9.60164e-05 9.56885e-05 9.5335e-05 9.48194e-05 9.47075e-05 9.42735e-05 9.36896e-05 9.3293e-05 9.3138e-05 9.28606e-05 9.25712e-05 9.25475e-05 9.18092e-05 9.08798e-05 9.08586e-05 9.05949e-05 8.99921e-05 8.95274e-05 8.91731e-05 8.90469e-05 8.83111e-05 8.774e-05 8.71845e-05 8.70884e-05 8.69316e-05 8.69962e-05 8.68079e-05 8.62112e-05 8.5765e-05 8.59066e-05 8.56022e-05 8.5392e-05 8.53058e-05 8.49568e-05 8.45074e-05 8.4476e-05 8.4289e-05 8.37752e-05 8.36267e-05 8.30102e-05 8.27922e-05\\
    scene0567_01    0.00241787 0.00181677 0.00144083 0.0012427 0.00107219 0.000979494 0.000877085 0.000803251 0.000748701 0.00069209 0.000645807 0.000604404 0.000570852 0.000538555 0.000511898 0.000489762 0.000465657 0.000446961 0.000427599 0.000412789 0.000396746 0.000383556 0.000373541 0.00036218 0.000352884 0.000342356 0.000331719 0.000324189 0.000315844 0.000308007 0.000300996 0.000293638 0.000289208 0.000284279 0.000278485 0.000273516 0.00026758 0.000264554 0.000258659 0.000254567 0.000249123 0.000245128 0.000241031 0.000237572 0.000234079 0.000230564 0.000225948 0.000222423 0.000218668 0.000214673 0.000211718 0.000208972 0.000206207 0.000203076 0.000201045 0.000198043 0.000195815 0.000193225 0.000191019 0.000188375 0.000185364 0.000183365 0.000181945 0.000179472 0.000177943 0.000176159 0.000174907 0.000173024 0.000171201 0.000170424 0.000168192 0.000166483 0.000165263 0.000162784 0.000161298 0.000159452 0.000157782 0.000156393 0.000154924 0.000153836 0.00015242 0.000151186 0.000149825 0.000148552 0.000147919 0.000146626 0.000145707 0.000145247 0.000144616 0.000143162 0.000142343 0.000141471 0.000141407 0.000140758 0.000139524 0.000139377 0.000138695 0.000137745 0.000137101 0.000136113 0.000135674 0.000134094 0.000133894 0.000132617 0.000131964 0.000130782 0.00013027 0.000129868 0.000128711 0.000127504 0.000127191 0.00012622 0.000125583 0.000124749 0.000124238 0.000123124 0.000122639 0.000121658 0.000120671 0.000120333 0.000119558 0.000119081 0.000118199 0.000117057 0.000116951 0.000115755 0.000114972 0.000114196 0.000113611 0.000112626 0.000112027 0.000111223 0.000110721 0.00011021 0.000109616 0.000108614 0.000107885 0.000107579 0.000107058 0.000106342 0.000105978 0.00010547 0.000104685 0.00010385 0.000103787 0.000102791 0.000102411 0.000101421 0.000100833 0.000100444 9.97847e-05 9.91399e-05 9.86865e-05 9.77394e-05 9.73314e-05 9.66394e-05 9.6172e-05 9.58624e-05 9.50863e-05 9.44191e-05 9.41157e-05 9.32985e-05 9.29646e-05 9.23768e-05 9.20869e-05 9.17958e-05 9.12627e-05 9.13135e-05 9.06655e-05 9.03501e-05 9.02258e-05 8.97052e-05 8.92771e-05 8.89704e-05 8.86822e-05 8.82149e-05 8.79872e-05 8.73161e-05 8.68083e-05 8.63866e-05 8.63055e-05 8.58827e-05 8.54364e-05 8.50309e-05 8.47531e-05 8.43409e-05 8.38298e-05 8.38009e-05 8.32568e-05 8.27559e-05 8.23973e-05 8.21904e-05 8.22705e-05 8.18546e-05 8.14653e-05 8.14735e-05 8.09395e-05 8.03833e-05\\
    }
    \datatable;
    \pgfplotstabletranspose[colnames from=cellvisitcount]\datatabletranspose{\datatable};

    \tikzstyle{styleplot}=[mark=none, very thick, opacity=0.6]

    \addplot +[styleplot, solid ] table [x=colnames, y=apt]          {\datatabletranspose}; \addlegendentry{apt}
    \addplot +[styleplot, solid ] table [x=colnames, y=offices]      {\datatabletranspose}; \addlegendentry{offices}
    \addplot +[styleplot, solid ] table [x=colnames, y=office0]      {\datatabletranspose}; \addlegendentry{office0}
    \addplot +[styleplot, solid ] table [x=colnames, y=office1]      {\datatabletranspose}; \addlegendentry{office1}
    \addplot +[styleplot, solid ] table [x=colnames, y=office3]      {\datatabletranspose}; \addlegendentry{office3}
    \addplot +[styleplot, solid ] table [x=colnames, y=scene0220_02] {\datatabletranspose}; \addlegendentry{scene0220\_02}
    \addplot +[styleplot, solid ] table [x=colnames, y=scene0271_01] {\datatabletranspose}; \addlegendentry{scene0271\_01}
    \addplot +[styleplot, solid ] table [x=colnames, y=scene0294_02] {\datatabletranspose}; \addlegendentry{scene0294\_02}
    \addplot +[styleplot, solid ] table [x=colnames, y=scene0451_05] {\datatabletranspose}; \addlegendentry{scene0451\_05}
    \addplot +[styleplot, dashed] table [x=colnames, y=scene0567_01] {\datatabletranspose}; \addlegendentry{scene0567\_01}

\end{axis}
\end{tikzpicture}
\caption{\addedSinceECCV{Convergence} of the average \addedSinceECCV{distance to shape} in proxy cells over time. %
Absolute value of the \addedSinceECCV{variation} of the average \addedSinceECCV{distance to shape} with relation to the number of depth samples accumulated within the proxy cell. %
\addedSinceECCV{At each new sample, we compute the absolute difference between the previous and current average values of distance to shape and average it over all cells in the scene.
We observe the fast convergence of the depth average after about 30 accumulated samples, where the change in distance to shape value falls below 0.5mm.
This shows that only a few seconds are needed for the statistics in the proxy to become stable.}}
\label{S_fig:avgincrement}
\end{figure*}

\autoref{S_fig:avgincrement} shows the fast convergence of the proxy statistics after about 30 accumulated samples, through the \addedSinceECCV{decreasing variation} over time of the average \addedSinceECCV{distance to shape}. %

\subsection{Timing Repartition}
\label{S_sec:experiments_timings}

\begin{figure*} %
\centering
    \includegraphics[width=0.8\textwidth]{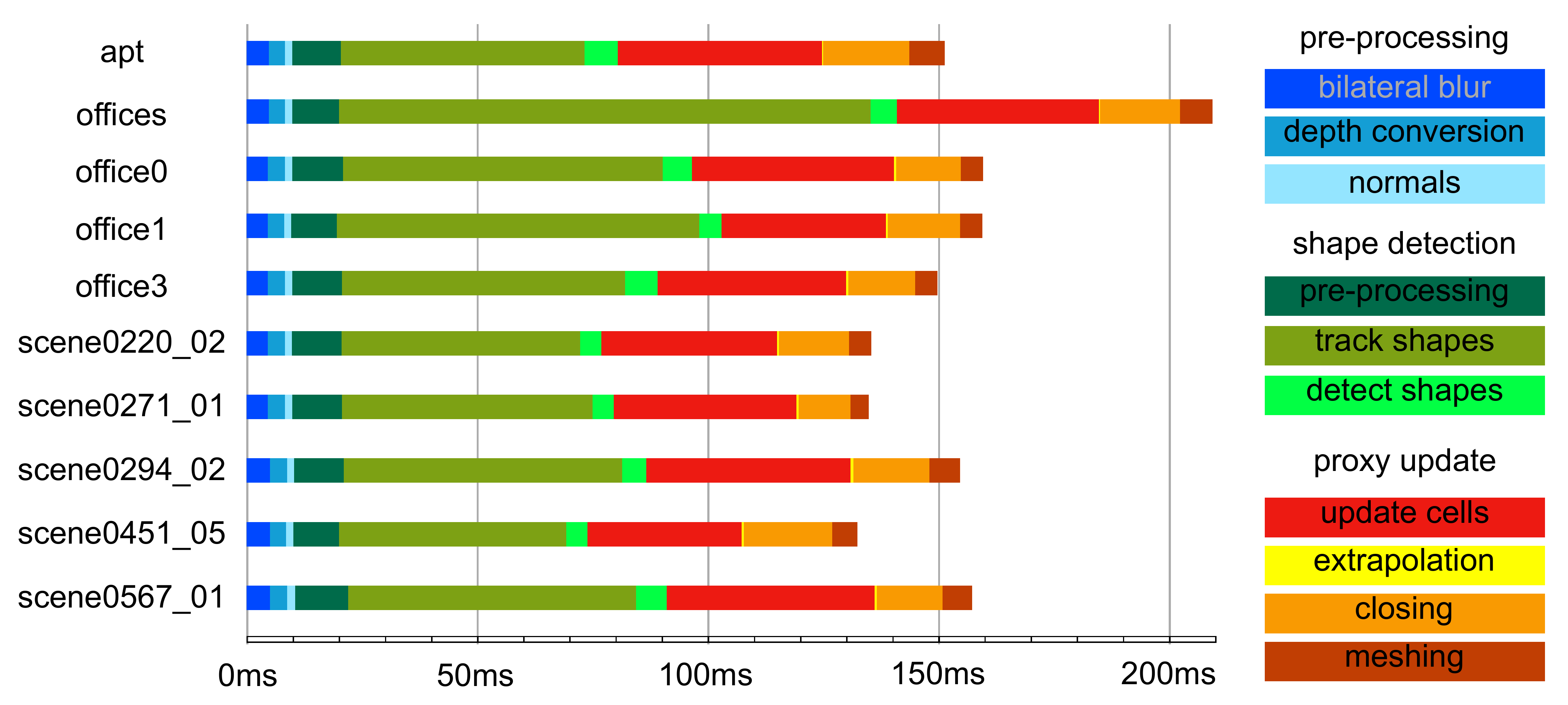}
\caption{Timing repartition for \addedSinceECCV{geometric} proxy generation, averaged over all frames of each scene.
\addedSinceECCV{Shape tracking is the longest processing with about 40\% of the total time, as it iterates over all previous proxies and depth image samples.
Proxy cell update is the second longest with about 30\% due to the iterations over cells and color points for each cell.
We can see that the tracking takes more time for the largest scene \emph{offices} (8518 frames), as its size implies more previous shapes to iterate over.}}
\label{S_fig:timings}
\end{figure*}

\autoref{S_fig:timings} shows the detailed repartition of the processing time \addedSinceECCV{for all steps of the proxy construction}.

\subsection{JPEG Export Metrics}
\label{S_sec:experiments_jpegexport}

\autoref{S_tab:compressionjpeg} reports compression metrics for offline export and storage of the proxies using \emph{JPEG} \cite{wallace1992jpeg}, \addedSinceECCV{detailed in \autoref{sec:experiments_processing_compression} of the paper}.

\begin{table}[ht]
\setlength{\tabcolsep}{7.3pt} %
\caption{Proxy JPEG export metrics} %
\label{S_tab:compressionjpeg}
\sffamily
{\centering
\begin{tabular}{ c  c c c }
    \multirow{2}{*}{scene}
    & \multicolumn{3}{ c }{\emph{JPEG} export}\\
    & compression ratio & PSNR (dB) & timing (ms)\\
    \hline
    apt & 8.09 & 33.8 & 41 \\
    offices & 9.22 & 33.8 & 71 \\
    office0 & 6.60 & 20.6 & 34 \\
    office1 & 6.70 & 34.7 & 41 \\
    office3 & 8.58 & 31.0 & 21 \\
    scene0220\_02  & 6.29 & 31.9 & 45 \\
    scene0271\_01 & 5.82 & 36.7 & 30 \\
    scene0294\_02 & 7.27 & 36.8 & 30 \\
    scene0451\_05 & 5.44 & 34.7 & 25 \\
    scene0567\_01 & 9.00 & 35.3 & 34 \\
    \hline
    \\
\end{tabular}}
\emph{The compression ratio is between the sizes of the \addedSinceECCV{full} raw proxy set and the exported \addedSinceECCV{JPEG image files}.
The \emph{peak signal-to-noise ratio (PSNR)} is computed using the average \emph{root mean square error (RMSE)} between raw depth points and \addedSinceECCV{their positions after loading proxies exported to JPEG}.
\addedSinceECCV{The reported timing is the total time required to export and load all scene proxies.}}
\end{table}

\section{RGB-D Data Consolidation}
\label{S_sec:consolidation}

\subsection{Close-up Comparison}
\label{S_sec:consolidation_reconscloseup}

\begin{figure*} %
\centering
    \includegraphics[width=\textwidth]{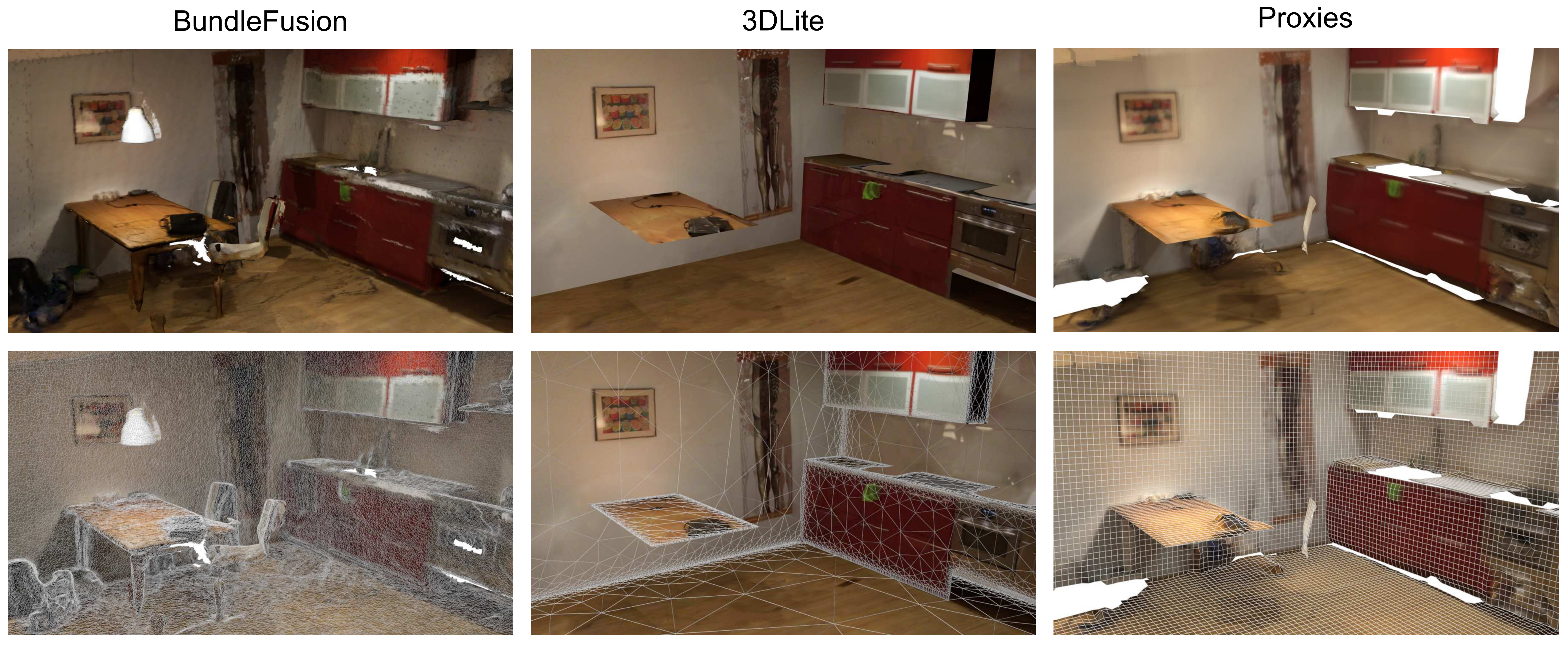}
\caption{\addedSinceECCV{Close-up views of reconstructions using our Proxies, \emph{BundleFusion} \cite{dai2017bundle} and \emph{3DLite} \cite{huang20173dlite}.
Top: We can notice the high detail but also high irregularity of the \emph{BundleFusion} mesh (left).
The \emph{3DLite} model (center) has sharp texture for high appearance quality, although no geometric details.
The proxy mesh (right) has a lower color resolution but sufficient to identify elements in the scene.
Its local information allows keeping geometric details smoothed out by \emph{3DLite}, such as the laptop on the table.
Bottom: The \emph{BundleFusion} mesh (left) is shown with only 30\% of the original geometry for better visualization,
but we can see that the high details imply a large amount of polygons.
\emph{3DLite} (center) is aware of the geometry and models large planar areas with large triangles, refining it at the limits of elements.
The regular grid of the proxies (right) is lightweight while storing accurate geometry at all locations of the shapes e.g., for the frame on the wall above the table.}}
\label{S_fig:consolidation_closeup}
\end{figure*}

\addedSinceECCV{\autoref{S_fig:consolidation_closeup} compares reconstructions with Proxies, \emph{BundleFusion} and \emph{3DLite} from a local, close-up point of view.
See \autoref{sec:consolidation_qualitative} of the paper for more details.}

\section{Reconstruction Regularization}
\label{S_sec:smoothpoisson}

\begin{figure*} %
\centering
    \includegraphics[width=\textwidth]{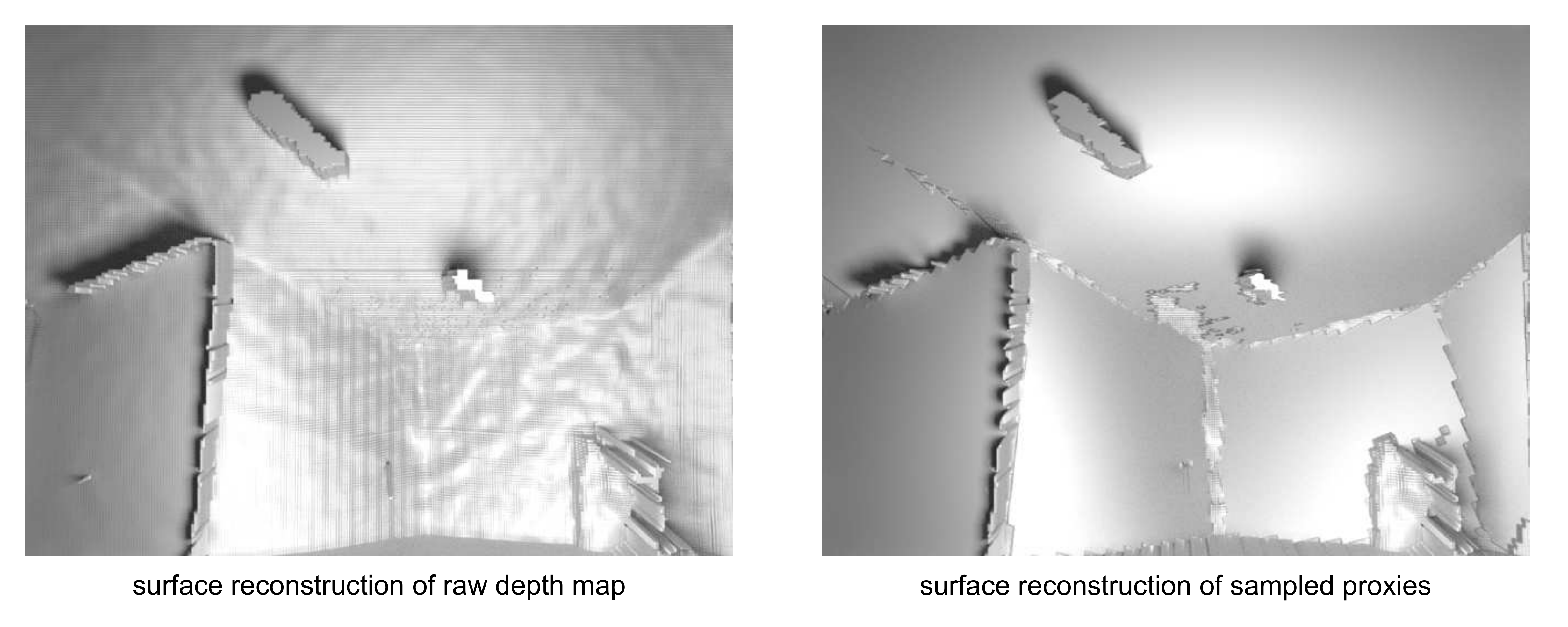}
\caption{\addedSinceECCV{Surface reconstruction on raw depth and proxy-filtered data.
This preliminary result shows how geometric proxies can be used as a smoothing operator in the context of mesh reconstruction.
Here, we apply \emph{Poisson surface reconstruction} \cite{kazhdan2006poisson} to the raw depth point cloud (left) and point cloud generated by sampling our proxies (right).
We can see that the planar elements (door on the left, ceiling at the top, and walls) are smoothly reconstructed
and do not exhibit the artifacts seen on the raw depth due to sensor limitations.}}
\label{S_fig:smoothupsamppoisson}
\end{figure*}

\addedSinceECCV{\autoref{S_fig:smoothupsamppoisson} shows preliminary results of surface reconstruction applied before and after proxy filtering, as detailed in \autoref{sec:conclusion_regularize} of the paper.}

\end{document}